\pdfoutput=1
\documentclass[11pt]{article}
\usepackage[english]{babel}
\usepackage{pdfpages}

\usepackage[final]{acl}

\usepackage{times}
\usepackage{latexsym}
\usepackage[T1]{fontenc}
\usepackage[utf8]{inputenc}

\usepackage{amssymb}
\usepackage{amsmath}
\usepackage{hyperref}
\usepackage{longtable}

\usepackage{microtype}

\usepackage{inconsolata}

\usepackage{graphicx}

\usepackage{subcaption}
\usepackage{placeins}

\usepackage{comment}
\usepackage{booktabs}
\usepackage{multirow}
\usepackage{array}

\title{Prompt, Translate, Fine-Tune, Re-Initialize, or Instruction-Tune?\\ Adapting LLMs for In-Context Learning in Low-Resource Languages}

\author{Christopher Toukmaji \\
  University of California, Irvine \\
  \texttt{ctoukmaj@uci.edu} \\\And
  Jeffrey Flanigan \\
  University of California, Santa Cruz \\
  \texttt{jmflanig@ucsc.edu} \\}

\usepackage{tcolorbox}

\pgfkeys{
  /pgfplots/error bars/error mark and bar options/.code={%
    \pgfkeysalso{
      /pgfplots/error bars/error mark options/.append={,#1},
      /pgfplots/every error bar/.append style={#1}
    }%
  }
}
\begin{document}

\maketitle

\begin{abstract}
    LLMs are typically trained in high-resource languages, and tasks in lower-resourced languages tend to underperform the higher-resource language counterparts for in-context learning. Despite the large body of work on prompting settings, it is still unclear how LLMs should be adapted cross-lingually specifically for in-context learning in the low-resource target languages. We perform a comprehensive study spanning five diverse target languages, three base LLMs, and seven downstream tasks spanning over 4,100 GPU training hours (9,900+ TFLOPs) across various adaptation techniques: few-shot prompting, translate-test, fine-tuning, embedding re-initialization, and instruction fine-tuning. Our results show that the few-shot prompting and translate-test settings tend to heavily outperform the gradient-based adaptation methods. To better understand this discrepancy, we design a novel metric, \emph{Valid Output Recall} (VOR), and analyze model outputs to empirically attribute the degradation of these trained models to catastrophic forgetting. To the extent of our knowledge, this is the largest study done on in-context learning for low-resource languages with respect to train compute and number of adaptation techniques considered. We make all our datasets and trained models available for public use.\footnote{\href{https://huggingface.co/collections/ChrisToukmaji/toukmaji-flanigan-gem25-684b85204462f36f765c4c0f}{https://huggingface.co/collections/ChrisToukmaji/toukmaji-flanigan-gem25}}
\end{abstract}

\section{Introduction}
\begin{figure}[t]
    \centering
    \includegraphics[width=\columnwidth]{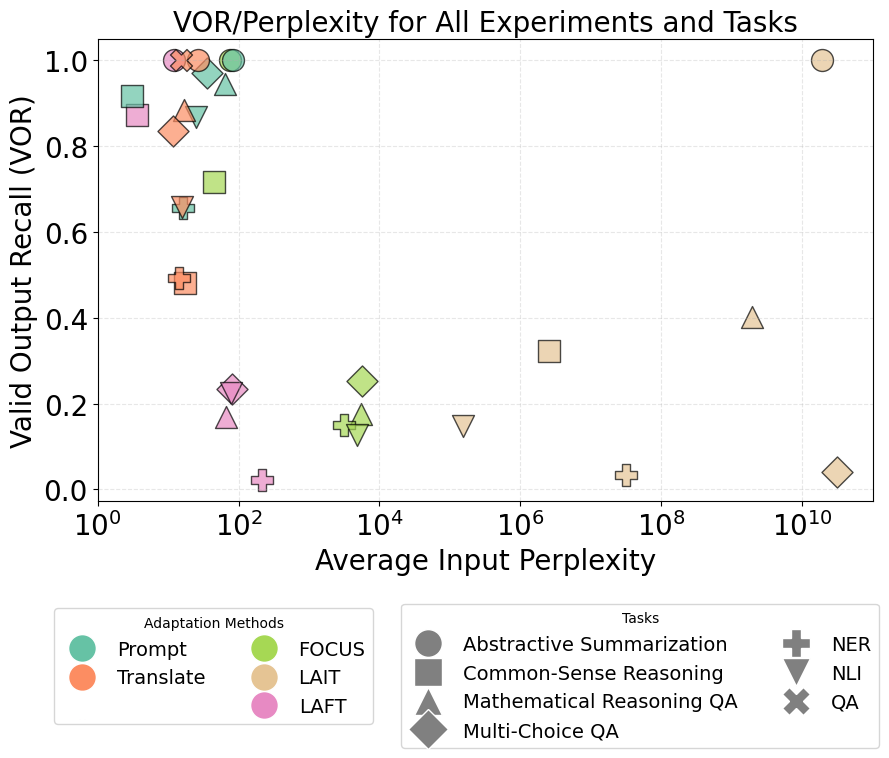}
    \caption{
    We report VOR scores (Valid Output Recall, the proportion of model outputs that follow the in-context labeling scheme) vs input perplexity for each adaptation method and task, averaged across target languages, and random seeds. Prompting-based methods (Prompt, Translate) demonstrate lower perplexity and higher VOR than gradient-based methods, suggesting that gradient-based methods suffer from catastrophic forgetting, degrading both linguistic ability and instruction-following alignment after training. Trained models lose the ability to learn in-context post-training while simultaneously performing worse on the target language.
    }
    \label{fig:VOR}
\end{figure}

\begin{table*}[h]
\centering
\scriptsize
\begin{tabular}{p{0.07\columnwidth} p{0.1\columnwidth} p{0.4\columnwidth} p{0.1\columnwidth} p{0.17\columnwidth} p{0.38\columnwidth} p{0.48\columnwidth}}
    \toprule
    \textbf{Lang.} & \textbf{Script} & \textbf{Family} & \textbf{Speakers} & \textbf{Word Order} & \textbf{Tasks Evaluated} & \textbf{Dataset Name} \\
    \midrule
    \multirow{4}{*}{\texttt{hau}} 
        & Latin & Afro-Asiatic-Chadic & 88M & SVO 
        & NER & MasakhaNER 
        \cite{adelani2021masakhaner} \\
        &  &  &  &  
        & Mathematical Reasoning QA & AfriMGSM \cite{adelani2024irokobenchnewbenchmarkafrican}  \\
        &  &  &  &  
        & NLI & AfriXNLI \cite{adelani2024irokobenchnewbenchmarkafrican}  \\
        &  &  &  &  
        & Abstractive Summarization & XL-Sum \cite{hasan-etal-2021-xlsum} \\
        &  &  &  &  
        & Multi-Choice QA & AfriMMLU \cite{adelani2024irokobenchnewbenchmarkafrican} \\
    \midrule
    \multirow{3}{*}{\texttt{lug}} 
        & Latin & Niger-Congo-Bantu & 11M & SVO 
        & NER & MasakhaNER \cite{adelani2021masakhaner} \\
        &  &  &  &  
        & Mathematical Reasoning QA & AfriMGSM \cite{adelani2024irokobenchnewbenchmarkafrican}  \\
        &  &  &  &  
        & NLI & AfriXNLI \cite{adelani2024irokobenchnewbenchmarkafrican}  \\
        &  &  &  &  
        & Multi-Choice QA & AfriMMLU \cite{adelani2024irokobenchnewbenchmarkafrican} \\
    \midrule
    \multirow{3}{*}{\texttt{kin}} 
        & Latin & Niger-Congo-Bantu & 15M & SVO 
        & NER & MasakhaNER \cite{adelani2021masakhaner} \\
        &  &  &  &  
        &  Mathematical Reasoning QA & AfriMGSM \cite{adelani2024irokobenchnewbenchmarkafrican}  \\
        &  &  &  &  
        & NLI & AfriXNLI \cite{adelani2024irokobenchnewbenchmarkafrican}  \\
        &  &  &  &  
        & Multi-Choice QA & AfriMMLU \cite{adelani2024irokobenchnewbenchmarkafrican} \\
    \midrule
    \multirow{2}{*}{\texttt{bur}} 
        & Burmese & Sino-Tibetan-Tibeto-Burman & 43M & SOV 
        & NER & Wiki-ANN \cite{pan-etal-2017-cross} \\
        &  &  &  &  
        & NLI & MyanmarXNLI \cite{Htet_MyanmarXNLI} \\
        &  &  &  &  
        & Abstractive Summarization & XL-Sum \cite{hasan-etal-2021-xlsum} \\
        &  &  &  &  
        & Common-Sense Reasoning & XStoryCloze \cite{lin2022fewshotlearningmultilinguallanguage} \\
    \midrule
    \multirow{4}{*}{\texttt{tha}} 
        & Thai & Tai-Kra-Dai & 69M & SVO 
        & NER & Wiki-ANN \cite{pan-etal-2017-cross} \\
        &  &  &  &  
        & Mathematical Reasoning QA & MGSM \cite{lmsarefewshotmultilingCOT} \\
         &  &  &  &  
        & NLI & XNLI \cite{conneau2018xnlievaluatingcrosslingualsentence} \\
        &  &  &  &  
        & Abstractive Summarization & XL-Sum \cite{hasan-etal-2021-xlsum} \\
        &  &  &  &  
        & Common-Sense Reasoning & XCOPA \cite{ponti-etal-2020-xcopa} \\
        &  &  &  &  
        & QA & XQUAD \cite{artetxe-etal-2020-cross} \\
    \bottomrule
\end{tabular}
\caption{The evaluated languages (ISO 639-2 code), written script, language family, number of speakers, word order typology, and the tasks/datasets we evaluate them on.}
\label{tab:language_tasks}
\end{table*}

% background of llm
    Large language models (LLMs) have been at the forefront of the advancements in Natural Language Processing (NLP), evidenced by state-of-the-art results on numerous benchmarks \cite{transformer, gpt3}. LLMs are pre-trained with large corpora of English text data, so the best LLMs are primarily monolingual and English-based, leaving other languages behind.
% what is the issue
    Performance for tasks in non-English languages tend to underperform the same task in English for LLMs \cite{ahuja2023megamultilingualevaluationgenerative, ahuja2024megaversebenchmarkinglargelanguage}. 
% why is this an issue
    Resource limitations prevent speakers of low-resource languages from participating in modern-day NLP since LLMs need considerable amounts of training data, and the most-capable LLMs perform poorly on low-resource languages compared to higher-resourced languages for in-context learning \cite{lai2023chatgpt, adelani2024comparingllmpromptingcrosslingual}. This exclusion is a particularly crucial issue, as most languages are low-resource, and these languages have billions of speakers \cite{magueresse2020lowresource}.
% why previous approaches dont wokr

    There have been several approaches to help make LLMs more multilingual. One approach involves pre-training an LLM from scratch on a non-English language \cite{martin-etal-2020-camembert, koto-etal-2020-indonesian-scratch, wilie-etal-2020-indonesian-scratch2, Polignano2019-italianscratch, canete2023spanish, kakwani-etal-2020-indicnlpsuite-indianLM, thapa2024developmentpretrainedtransformerbasedmodels}, but this approach assumes access to a sufficiently-large corpus of text and significant computational resources. % even if you have data, doesnt generalize 
    Another prevalent approach is multilingual LLMs, in which an LLM is pre-trained on many different languages \cite{XLM, mbert, curse-multilinguality, mbart, xue2021mt5, afriberta, lin2022fewshotlearningmultilinguallanguage}. However, as more languages are introduced, the monolingual and cross-lingual performance deteriorates \cite{curse-multilinguality} with low-resource languages being far more vulnerable \cite{wu-dredze-2020-languages}. As a result, a large focus in the area of cross-lingual transfer has been attempting to retain the strong performance of primarily-monolingual LLMs for other non-English languages.  However, these results display that the best approach fluctuates across base models, languages, and tasks \cite{ahuja2023megamultilingualevaluationgenerative}.

    We perform a systematic evaluation of cross-lingual transfer approaches specifically for in-context learning to identify patterns for optimal transfer settings. To the extent of our knowledge, this is the largest study (with respect to TFLOPs and GPU training hours) on cross-lingual transfer for in-context learning in low-resource languages spanning three base LLMs, five low-resource target languages, five adaptation methods, and seven NLP tasks.  Our results show that the prompt and translate settings tend to heavily outperform the gradient-based adaptation methods. To better understand this discrepancy, we design \emph{Valid Output Recall} (VOR), a novel metric, and analyze model outputs to empirically attribute the degradation of these trained models to catastrophic forgetting.

\begin{figure*}[p]
    \centering
    \begin{subfigure}[b]{\linewidth}
        \centering
        \includegraphics[width=\linewidth]{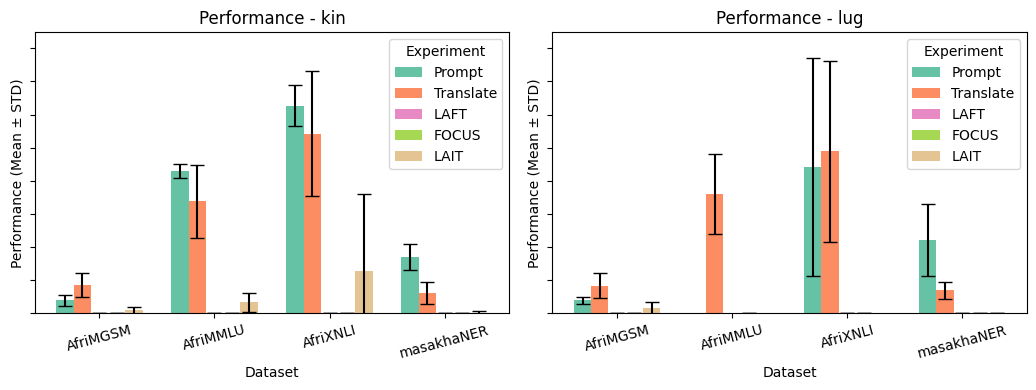}
    \end{subfigure}
    \begin{subfigure}[b]{\linewidth}
        \centering
        \includegraphics[width=0.6\linewidth]{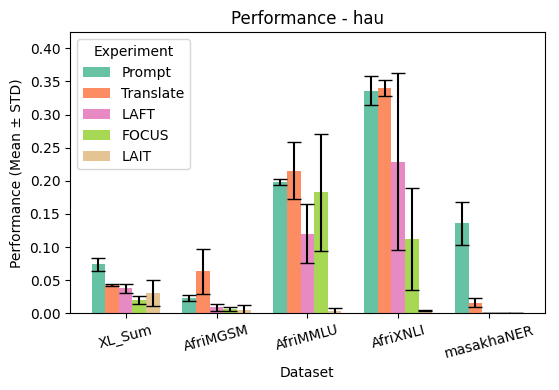}
        \label{fig:results-1}
    \end{subfigure}
    \vspace{0.5cm} % Adjust vertical spacing as needed
    \begin{subfigure}[b]{\linewidth}
        \centering
        \includegraphics[width=\linewidth]{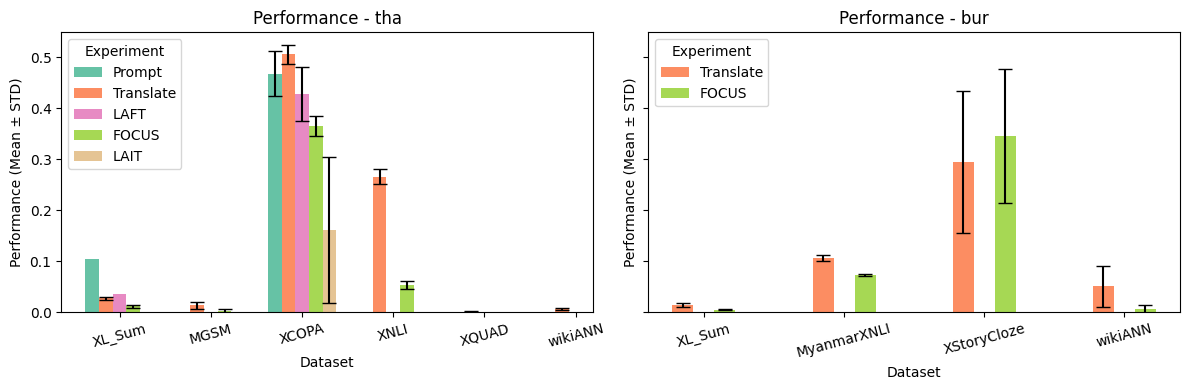}
        \label{fig:results-2}
    \end{subfigure}
    \caption{Few-shot downstream task performance across various adaptation methods for all five languages evaluated: Hausa (\texttt{hau}), Luganda (\texttt{lug}), Kinyarwanda (\texttt{kin}), Thai (\texttt{tha}) and Burmese (\texttt{bur}), averaged over three random seeds. We evaluate five adaptation methods: prompting-based methods (Prompt, Translate) and gradient-based methods (Language-Adaptive Fine-Tuning (LAFT), FOCUS (embedding re-initialization + LAFT), and Language-Adaptive Instruction Tuning (LAIT)). We find that prompting-based methods consistently outperform gradient-based methods across all tasks.}
    \label{fig:combined-results}
\end{figure*}

\section{Related Work}

The work of \citet{tejaswi2024exploringdesignchoicesbuilding} is the most similar to ours. This work evaluates multilingual adaptation of LLMs for in-context learning with an emphasized study on vocabulary expansion and embedding re-initialization strategies. This study finds that that vocabulary expansion and embedding re-initialization can help bridge the gap between the performance of English and non-English languages in LLMs. Our work differs from this in that embedding re-initialization is just one of the adaptation methods that we evaluate in our study. 

\citet{ahuja2023megamultilingualevaluationgenerative} perform a study that evaluates on a subset of our adaptation methods - namely, translate-test and few-shot prompting. The study finds that the translate-test adaptation method outperforms few-shot prompting in most languages and tasks. This work differs from ours in that the study only considers prompt-based adaptation methods and no gradient-based approaches, like ours does. \citet{ahuja2024megaversebenchmarkinglargelanguage} conduct an analysis on few-shot prompting across many language models, tasks, and languages, but do not consider any other adaptation methods.

Others have benchmarked adaptation methods, but they differ from our study in that our main focus is on several low-resource languages \cite{asai2023buffetbenchmarkinglargelanguage,toraman-2024-adapting, wang-etal-2025-language}

There is a multitude of work in cross-lingual transfer, but only the papers above have a similar emphasis of benchmarking adaptation methods for in-context learning. 
Other primary lines of work in cross-lingual transfer from monolingual LLMs include 1) performing further training with the target language \cite{african-langauge-adaptive-ft,joshi2024adaptingmultilingualllmslowresource, doshi2024worrydatabuildingpretrained, razumovskaia2024analyzingadaptinglargelanguage, sani2025extendingllmsnewlanguages}, 2) modifying the embedding matrix and vocabulary to better fit the target language \cite{recyclegpt2, dobler-de-melo-2023-focus, remy2023tiktotoktranslatinglanguagemodels,gosal2024bilingualadaptationmonolingualfoundation, cui2024efficienteffectivetextencoding, mundra2024empiricalcomparisonvocabularyexpansion, yamaguchi2024effectivelyexpandvocabularyllms, pham2024unibridgeunifiedapproachcrosslingual, da-dalt-etal-2024-flor, vres2024generativemodellessresourcedlanguage, yamaguchi2024empiricalstudycrosslingualvocabulary} or 3) instruction-tuning with the target language \cite{kuulmets2024teachingllamanewlanguage} or cross-lingually \cite{chen2024translationfusionimproveszeroshot, ranaldi2023empowering, ranaldi-pucci-2023-english}. These generalized cross-lingual transfer techniques are evaluated in our study.

\section{Methods}
        We use three different multi-billion parameter base LLMs - LLaMa 2 7B \cite{touvron2023llama2openfoundation}, MPT-7B \cite{databricksIntroducingMPT7B}, and Phi-2 \cite{microsoftPhi2Surprising} - for in-context learning in five diverse low-resource languages. We opt to use these models since they are primarily-monolingual, open-source, and are capable of in-context learning. We use English as our source language and evaluate on a set of five diverse low-resource target languages: Hausa (\texttt{hau}), Kinyarwanda (\texttt{kin}),  Luganda (\texttt{lug}), Burmese (\texttt{bur}), and Thai (\texttt{tha}).

            \subsection{Evaluation Setting}
            \label{dataset-eval}
            We aim to evaluate scenarios where a target language speaker uses an LLM to perform a downstream task in that language. Accordingly, we only consider tasks and datasets where the task instance is in the target language. Not every task is evaluated in every language because we do not have datasets for all these tasks in each language. 

            During evaluation, we form our few-shot prompts with a random sample without replacement of the training split for the evaluation datasets outlined in \autoref{tab:language_tasks}. For all settings, each shot is prepended with a machine-translated description of the task in the target language. We use the maximum number of shots that fit within the context length for each dataset. The reported results are on the test split of the dataset for that language, and we conduct three samples with random seeds and average the performance across the test split. Some experiments were omitted due to context window or memory limitations (see \hyperref[unperformed]{Appendix~\ref*{unperformed}} for details).
        
            We emphasize that no task-specific fine-tuning is done at any point in any of our experiments. Our core research questions aims to answer how to transfer LLMs to new languages while remaining as general-purpose task solvers. 

        \subsection{Datasets and Metrics}
        The datasets we evaluate on for each langauge are given in \autoref{tab:language_tasks}. We report F1-score for MasakhaNER and WikiANN, ROUGE-L for XL-Sum, and accuracy for AfriMGSM, AfriXNLI, AfriMMLU, MGSM, XCOPA, and XStoryCloze. MasakhaNER and WikiANN are language-specific datasets, whereas the others are either evaluated in a single language or are parallel.

\section{Experiments}

        We evaluate the following five methods for adapting an LLM trained in a source language for prompting with a target language.

        \paragraph{{Few-Shot Prompting (Prompt)}} We prompt the LLM with the few-shot prompt in the target language and evaluate the completion. This method requires no translation, nor any gradient updates.
        
        \paragraph{{Translate-test (Translate)}} We first machine-translate the few-shot prompt from the target language to the source language. Next, the LLM is prompted in the source language. Then, the output is translated from the source language back to the target language. We use NLLB-200 3.3B \cite{nllbteam2022languageleftbehindscaling} for both translation directions. This method does not require any gradient updates \cite{hu2020xtrememassivelymultilingualmultitask}.
        
        \paragraph{{Language-Adaptive Fine-Tuning (LAFT)}}      Starting with the original LLM, we further fine-tune the LLM on a corpus of tokens in the target language using the original pre-training objective. Then, we prompt the LLM.

        \paragraph{{Vocabulary and Embedding Re-initialization (FOCUS)}}   
        Following \citet{dobler-de-melo-2023-focus}, we perform the FOCUS method; we train a new tokenizer in the target language, then use pre-trained static embeddings \footnote{\href{https://fasttext.cc/docs/en/pretrained-vectors.html}{https://fasttext.cc/docs/en/pretrained-vectors.html}} in the target language to re-initialize semantically-similar overlapping tokens in the embedding matrix of the base LLM, and thereafter perform LAFT on the LLM. Then, we prompt the LLM.

        \paragraph{{Language-Adaptive Instruction Tuning (LAIT)}}
        We machine-translate an Instruction Tuning dataset from the source language to the target language, then we perform instruction fine-tuning on the translated dataset. Then, we prompt the LLM.

\section{Results}

The results in \autoref{fig:combined-results} display few-shot prompting and translate-test adaptation methods surprisingly tend to heavily outperform the gradient-based adaptation methods. Below, we provide an empirical analysis of the LLMs' outputs, and show this disparity can be attributed to catastrophic forgetting \cite{MCCLOSKEY1989109}, which can occur in LLMs during continued training \cite{luo2025empiricalstudycatastrophicforgetting}. 
%instruction-followed label proportion

In order to determine whether the performance degradation is attributed to insufficient knowledge of the target language or to task forgetting, we design \emph{Valid Output Recall} (VOR), a metric to quantify an LLM's ability to instruction-follow labels in-context. VOR is the proportion of LLM outputs of a test set that follow the same labeling scheme that was instructed and provided in-context. For example, in a binary-classification task instance where an LLM is instructed to output a label $L \in \{0,1\}$ for test instance $i$ in a test dataset with size $N$ and the LLM output $\hat{y}_i$, then $VOR = \frac{1}{N} \sum\limits_{i=1}^{N} \mathbb{I}(\hat{y}_i \in L)$. The VOR is compared with the perplexity of the inputs. Intuitively, this isolates the evaluation of an LLM's task alignment and instruction-following ability from its linguistic ability. 

In \autoref{fig:VOR}, we observe that gradient-based adaptation methods have both higher perplexities and lower VOR compared to few-shot prompting. These results empirically verify that the trained models are losing the ability to learn in-context post-training while simultaneously performing worse on the target language. This suggests that the gradient-based methods in our training environment suffer from catastrophic forgetting since both linguistic knowledge and task alignment deteriorate.

\section{Conclusion}

This work provides the largest comprehensive study on adapting primarily-English LLMs to low-resource languages for in-context learning. Five adaptation methods are evaluated across three base LLMs using five diverse target languages on seven downstream tasks. Few-shot prompting and translate-test worked the best in nearly all cases, but there is no trend between which of the two works best. We design a novel metric, \emph{Valid Output Recall} (VOR), and provide an empirical analysis on LLM outputs to show that models adapted with gradient-based  methods degraded due to catastrophic forgetting.

\section*{Future Work}
In this work, we experiment with five diverse low-resource languages, but there are other low-resource languages that are also in need of more research. We leave this as future work, and we hope our work will help inspire research for other low-resource languages. 

We used two training-free adaptation methods: few-shot prompting and translate-test. There are other training-free prompting methods such as varying design templates or demonstration selections which we leave as future work. Given that training-free adaptation methods produced the best results in our paper, we are optimistic for future work in this direction, and believe our findings provide a strong motivation for further research into training-free adaptation approaches.

\section*{Limitations}
One limitation of our approach is that the translate-test setting hinges on an NMT model which could introduce translation errors and, in turn, affect performance. While this is a general issue with translation-based methods, future improvements in NMT quality could help reduce this effect.

\section*{Potential Risks}

We do not anticipate any potential risks with respect to ethical or social impacts from our work. However, since a component of our contributions is the open-sourcing of the trained models, we acknowledge that LLMs are capable of generating text that could be harmful \cite{gehman2020realtoxicitypromptsevaluatingneuraltoxic} or non-factual \cite{Huang_2025}.

\section*{Acknowledgments}
We thank Brendan King, Changmao Li, Chris Liu, Brian Mak, Nilay Patel, Geetanjali Rakshit, Rongwen Zhao, Zekun Zhao, Giridhar Vadhul, Ian Lane, and Amita Misra for their insightful feedback on earlier versions of this work. We would also like to thank the anonymous reviewers and area chairs for their detailed and helpful feedback.

This work used resources available through the National Research Platform (NRP) at the University of California, San Diego. NRP has been developed, and is supported in part, by funding from National Science Foundation, from awards 1730158, 1540112, 1541349, 1826967, 2112167, 2100237, and 2120019, as well as additional funding from community partners.

\bibliography{mybib}

\newpage
\newpage

\clearpage
\section*{Appendix}

\setcounter{subsection}{0}
\renewcommand{\thesubsection}{\Alph{subsection}}

\label{Appendix}

    \subsection{Training Details}
    Each model was trained using 1 NVIDIA A100-SXM4 80GB. For computational efficiency, we use DeepSpeed's Zero Redundancy Optimizer (ZeRO) at stage 3 \cite{deepspeedDeepSpeedZeRO3}, BF16 mixed precision training \cite{micikevicius2018mixedprecisiontraining}, a block size of 1024, and a 32-bit paged AdamW optimizer \cite{bitsandbytes}.

    \subsubsection{Hyperparameters}
    We use the standard practice of selecting the maximum batch size that fits within GPU memory. This results in a batch size of 1 for LAFT and FOCUS training, and a batch size of 2 for LAIT training.

    Each of the adaptation methods that required training (LAFT, FOCUS, LAIT) were trained for 6 epochs. We keep checkpoints after each epoch, and the model checkpoint with the lowest loss on the validation set is kept. We initialize training with the following hyperparameters taken from the LLaMa-2 paper \cite{touvron2023llama2openfoundation}: the AdamW optimizer \cite{loshchilov2019decoupledweightdecayregularization} with $\beta_1 = 0.9, \beta_2=0.95, \epsilon=10^{-5}$,  a learning rate of $3e^{-4}$ with a cosine scheduler warm-up of 2000 steps, $0.1$ weight decay, and gradient clipping at of $1$. We report the epoch of the best-performing checkpoint in \autoref{compute}.

    \subsection{Inference Details}
    Based on cluster availability, we use 1 of the following GPUs for inference: NVIDIA L40 (48GB), NVIDIA A4000 (16 GB), NVIDIA GeForce RTX 3090 (24GB), NVIDIA GeForce RTX 4090 (24GB), NVIDIA A10 (24GB), NVIDIA L4 (24GB). We use the standard practice of selecting the maximum batch size that fits within GPU memory. This value is one of $[4, 8, 16, 32]$ and is automatically calculated based on the size of the test instances and the GPU memory from whichever GPU the job is assigned. \\ \\ \\ \newline \newline \newline \newline
    
    \subsection{Budget}
    We report training runtime and TFLOPs for each trained model after all 6 epochs in \autoref{compute}.

    \begin{table}[h]
    \begin{scriptsize}
        \begin{tabular}{p{0.15\columnwidth}p{0.1\columnwidth}p{0.1\columnwidth}p{0.1\columnwidth}p{0.2\columnwidth}p{0.1\columnwidth}}
            \toprule
            Method & Lang. & Model & Best Epoch & Train Runtime (hours) & TFLOPs \\
            \midrule
            FOCUS & \texttt{bur} & llama & 5 & 206 & 240 \\
            FOCUS & \texttt{bur} & mpt & 5 & 160 & 240 \\
            FOCUS & \texttt{bur} & phi & 5 & 28 & 816 \\
            FOCUS & \texttt{hau} & llama & 5 & 234 & 240 \\
            FOCUS & \texttt{hau} & mpt & 3 & 167 & 240 \\
            FOCUS & \texttt{hau} & phi & 5 & 27 & 816 \\
            FOCUS & \texttt{kin} & llama & 4 & 87 & 116 \\
            FOCUS & \texttt{kin} & mpt & 4 & 98 & 105 \\
            FOCUS & \texttt{kin} & phi & 1 & 12 & 358 \\
            FOCUS & \texttt{lug} & llama & 4 & 6 & 7 \\
            FOCUS & \texttt{lug} & mpt & 4 & 17 & 7 \\
            FOCUS & \texttt{lug} & phi & 4 & 1 & 23 \\
            FOCUS & \texttt{tha} & mpt & 5 & 197 & 240 \\
            FOCUS & \texttt{tha} & phi & 5 & 32 & 816 \\
            LAFT & \texttt{bur} & llama & 5 & 217 & 240 \\
            LAFT & \texttt{bur} & mpt & 5 & 208 & 240 \\
            LAFT & \texttt{bur} & phi & 5 & 113 & 816 \\
            LAFT & \texttt{hau} & llama & 4 & 256 & 240 \\
            LAFT & \texttt{hau} & mpt & 5 & 267 & 240 \\
            LAFT & \texttt{hau} & phi & 5 & 111 & 816 \\
            LAFT & \texttt{kin} & llama & 4 & 219 & 227 \\
            LAFT & \texttt{kin} & mpt & 4 & 221 & 211 \\
            LAFT & \texttt{kin} & phi & 5 & 102 & 740 \\
            LAFT & \texttt{lug} & llama & 3 & 17 & 16 \\
            LAFT & \texttt{lug} & mpt & 3 & 12 & 15 \\
            LAFT & \texttt{lug} & phi & 4 & 7 & 52 \\
            LAFT & \texttt{tha} & llama & 4 & 200 & 240 \\
            LAFT & \texttt{tha} & mpt & 5 & 218 & 240 \\
            LAFT & \texttt{tha} & phi & 5 & 119 & 816 \\
            LAIT & \texttt{bur} & llama & 6 & 30 & 36 \\
            LAIT & \texttt{bur} & mpt & 6 & 119 & 36 \\
            LAIT & \texttt{bur} & phi & 6 & 3 & 142 \\
            LAIT & \texttt{hau} & llama & 3 & 20 & 10 \\
            LAIT & \texttt{hau} & mpt & 6 & 67 & 10 \\
            LAIT & \texttt{hau} & phi & 4 & 3 & 35 \\
            LAIT & \texttt{kin} & llama & 4 & 22 & 10 \\
            LAIT & \texttt{kin} & mpt & 6 & 65 & 9 \\
            LAIT & \texttt{kin} & phi & 4 & 3 & 32 \\
            LAIT & \texttt{lug} & llama & 6 & 20 & 12 \\
            LAIT & \texttt{lug} & mpt & 6 & 66 & 11 \\
            LAIT & \texttt{lug} & phi & 4 & 3 & 39 \\
            LAIT & \texttt{tha} & llama & 4 & 28 & 20 \\
            LAIT & \texttt{tha} & mpt & 5 & 97 & 21 \\
            LAIT & \texttt{tha} & phi & 4 & 3 & 107 \\
            \midrule
            \textbf{Total} & - & - & - & \textbf{4108} & \textbf{9943} \\
            \bottomrule
            \end{tabular}
    \end{scriptsize}
     \caption{Used training checkpoint, final training run-time in hours, and Tera Floating Point Operations  (TFLOPs) for every trained model}
     \label{compute}
    \end{table}

            \subsection{Training Datasets}

            \subsubsection{LAFT and FOCUS}
            
            We use the following language-specific fine-tuning corpora for LAFT and FOCUS.  For Burmese, Hausa, and Thai, we use a subset of mC4 \cite{xue2021mt5}, a multilingual variant of the C4 pre-training corpus \cite{raffel2020exploringc4}. For Kinyarwanda and Luganda, we use a subset of CommonVoice \cite{commonvoice_ardila2020} since an mC4 split doesn't exist for these languages. For all LAFT and FOCUS experiments, we train on 25M tokens and use the provided evaluation set for validation.  
            \label{dataset-fine-tune}

            \subsubsection{LAIT}
            We use a professional neural machine translation system \footnote{\href{https://cloud.google.com/translate/docs/reference/rest}{https://cloud.google.com/translate/docs/reference/rest}} to translate a random sample of 5,000 instruction-following examples from the Alpaca dataset \cite{alpaca}. We translate the same 5,000 instruction-following examples from English to each of the target languages, and we release the translated parallel instruction-following datasets on the HuggingFace dataset hub.\footnote{\href{https://huggingface.co/collections/ChrisToukmaji/toukmaji-flanigan-gem25-684b85204462f36f765c4c0f}{https://huggingface.co/collections/ChrisToukmaji/toukmaji-flanigan-gem25}} We designate $85\%$ of the examples for training and the remaining $15\%$ for validation.

    \subsection{Scientific Artifact Licenses}
    Below, we outline the scientific artifacts used (base models, training datasets, evaluation datasets) and the respective licenses. 
    \begin{table}[htbp]
    \centering
    \small
    \begin{tabular}{p{0.425\columnwidth}p{0.425\columnwidth}}
    \toprule
    \textbf{Artifact}  & \textbf{License} \\ \midrule
    LLaMa-2 7B & \textsc{llama2} \\
    MPT 7B & \textsc{apache-2.0} \\
    Phi-2 & \textsc{mit} \\
    NLLB-200 3.3B & \textsc{cc-by-na-4.0} \\
    \midrule
    mc4 & \textsc{odc-by} \\
    CommonVoice & \textsc{CC-0} \\
    \midrule
    AfriMGSM & \textsc{apache-2.0} \\
    AfriMMLU & \textsc{apache-2.0} \\
    AfriXNLI & \textsc{apache-2.0} \\
    MasakhaNER & \textsc{cc-by-nc-4.0} \\
    MyanmarXNLI & \textsc{cc-by-nc-4.0} \\
    MGSM & \textsc{mit} \\
    Wiki-ANN & \textsc{cc-0} \\
    XCOPA & \textsc{cc-by-4.0} \\
    XNLI & \textsc{cc-by-nc-4.0} \\
    XL-Sum & \textsc{cc-by-nc-sa-4.0} \\
    XQUAD & \textsc{cc-by-sa 4.0} \\
    XStoryCloze & \textsc{cc-by-sa-4.0} \\
    \bottomrule
    \end{tabular}
    \caption{Licenses for base models, training datasets, and evaluation datasets}
    \end{table}

    \subsection{Dataset Splits}

    \begin{table}[htbp]
    \centering
    \small
    \begin{tabular}{p{0.35\columnwidth}p{0.15\columnwidth}p{0.15\columnwidth}p{0.15\columnwidth}}
    \toprule
    \textbf{Dataset and Lang.} & train & eval & test  \\ \midrule
    AfriMGSM (all) & 8 & -  & 250 \\
    AfriMMLU (all) & - & 83 & 500 \\
    AfriXNLI (all) & - & 450 & 600 \\
    MasakhaNER (\texttt{hau}) & 1903 & 272 & 545  \\
    MasakhaNER (\texttt{kin}) & 2110 & 301 & 604  \\
    MasakhaNER (\texttt{lug}) & 2003 & 200 & 401  \\
    MyanmarXNLI & 392,702 & 2,490 & 5,010  \\
    MGSM & 8 & - & 250  \\
    Wiki-ANN (\texttt{bur}) & 100 & 100 & 100  \\
    Wiki-ANN (\texttt{tha}) & 20,000 & 10,000 & 10,000 \\
    XCOPA & - & 100 & 500 \\
    XNLI & 392,702 & 2,490 & 5,010  \\
    XL-Sum (\texttt{bur}) & 4,569 & 570 & 570 \\
    XL-Sum (\texttt{hau}) & 6,418 & 802 & 802\\
    XL-Sum (\texttt{tha}) & 6,616 & 826 & 826 \\
    XQUAD & - & 1,190 & - \\
    XStoryCloze & 361 & - & 1,511 \\
    \bottomrule
    \end{tabular}
    \caption{Evaluation dataset sizes for training, validation, and test datasets}
    \label{splits}
    \end{table}
    We outline the size of the train and test sets in \autoref{splits}. To form the few-shot prompt, we randomly sample from the training set, or the validation set if there is no train split. We report results on the test split. 
    
    XQUAD does not natively have a train/val/test split, so we use 10\% of the data for our `train' split and the remaining 90\% as the `test' split. We use the same split for all experiments.

    \subsection{Prompt Selection}
    Our block size is 1024, and we allocate 75\% of the block size (768 tokens) to context and 25\% of the block size (256 tokens) for the completion. In order to determine which train/eval instances to put in context, we perform the following steps. First, for every evaluation dataset, we find the largest instance in the set (in terms of tokens). In the worst case, this determines how many tokens are left in context for the completed exemplars/shots (i.e. if the largest test instance for a given dataset is 100 tokens, we must fit the completed exemplars within 668 tokens). Then, we randomly sample from the train/eval sets to try to get $k$ shots to fit within the remainder of the context window, where $k$ is the desired number of shots in-context. We maximize $k$ and stop sampling after 20 attempts. If we cannot fit even a single exemplar ($k=1$) after 20 tries, we are unable to perform inference for this experiment (see \autoref{tab:fullresults2}, \autoref{tab:fullresults1}, and Appendix \ref{unperformed} for a list and discussion of such instances). After performing these steps, we ended with a value of $k=1$ for all reported experiments, except for NLI tasks where we use a value of $k=3$.
    
    In order to perform to perform NLI faithfully, $k=3$ is the minimum value of shots to put into context since there needs to be one exemplar for each NLI label. When sampling from the train set in NLI experiments, we enforce a constraint that there must be one exemplar for each NLI label. The order of the NLI exemplars is randomized. 

    For NER tasks, we enforce a constraint that the exemplar in context must have at least one named-entity. All other tasks have no constrains on training data contents sampled for in-context learning.

    \subsection{Answer Extraction}

    We use the same cleaning procedure as outlined by \citet{touvron2023llama} for Question-Answering tasks, in which the answer is extracted from the generation by only considering content before the first line break, or the final dot/comma. For Mathematical-Reasoning QA, we extract the final space-separated integer since the output generation is Chain-of-Thought. For Multi-Choice QA, NLI, and Common-Sense Reasoning, we extract the first instance of the label set (\{A,B,C,D\} for Multi-Choice QA, \{0,1,2\} for NLI, and \{1,2\} for Common-Sense Reasoning). For Abstractive Summarization, we strip new line tokens. For NER, we strip out text outside the first occurrence of an opening and closing bracket, as implied by the label format in-context. The content within the brackets is filtered by only considering entity pairs with both opening and closing parentheses.

    We utilize these label sets and answer extraction methods when calculating VOR. Generation tasks like abstractive summarization are free-form and do not have to adhere to strict formatting which explains why the VOR scores are near perfect for generation tasks, but much smaller for tasks with strict required outputs (i.e. NLI). 

    As VOR is a recall-oriented metric, instances without an extracted answer following the pre-processing steps are treated as incorrect, whereas instances with any extracted answer, regardless of its semantic correctness, are treated as correct.

    \subsection{Unperformed Experiments}
    As outlined in \autoref{tab:fullresults2} and \autoref{tab:fullresults1}, a few experiments were infeasible to run. The FOCUS training task for \texttt{tha} with the LLaMa-2-7B model was infeasible to train due to memory constraints (more details below). The remainder of the excluded tasks were infeasible because they were unable to fit within the partition of the block size allocated for context.
    
    The FOCUS task for \texttt{tha} with the LLaMa-2-7B model required over 2TB of RAM to train a new tokenizer which far exceeded the 1TB RAM limits imposed on us from our compute cluster resource manager. We attempted to bypass this hurdle by renting a higher-capacity machine (with 2TB of RAM) from a popular cloud compute provider, but we were still unable to train the new tokenizer as it still exceeded the available RAM. Our study aims to emulate a compute-constrained environment and continuing to scale such an experiment to these increased levels would be in opposition to our objective. During debugging, we isolated the RAM issue as specific to the combination of the size of the mC4 Thai training split with the LLaMa-2 tokenizer. 
    \label{unperformed}

\begin{table*}[]
        \tiny
        \centering
        \caption{Few-shot downstream task performance in each training setting for Hausa (\texttt{hau}), Luganda (\texttt{lug}), and Kinyarwanda (\texttt{kin}) averaged over 3 runs for all models. We report F1-score for MasakhaNER, ROUGE-L for XL-Sum, and accuracy for AfriMGSM and AfriXNLI. The MasakhaNER dataset is specific to each language, but AfriMGSM and AfriXNLI are parallel.}
        \label{tab:fullresults2}
        \begin{tabular}{p{0.78cm}p{0.67cm}p{0.8cm}p{0.8cm}p{0.6cm}p{0.8cm}|p{0.75cm}p{0.7cm}p{0.7cm}p{0.75cm}|p{0.75cm}p{0.7cm}p{0.67cm}p{0.65cm}}
            \toprule \\
            \multicolumn{13}{c}{{\emph{\textbf{LLaMa-2 7B}}}} \\
            \cmidrule{2-14} \\
            \multirow{2}{*}{\textit{Experiment}} & \multicolumn{5}{c}{\textbf{hau}} & \multicolumn{4}{c}{\textbf{lug}} & \multicolumn{4}{c}{\textbf{kin}} \\
            \cmidrule(lr){2-6} \cmidrule(lr){7-10} \cmidrule(lr){11-14}
            & \textit{XL-Sum} & \textit{MasakhaNER} & \textit{AfriMGSM} & \textit{AfriXNLI} & \textit{AfriMMLU} & \textit{MasakhaNER} & \textit{AfriMGSM} & \textit{AfriXNLI} & \textit{AfriMMLU} & \textit{MasakhaNER} & \textit{AfriMGSM} & \textit{AfriXNLI} & \textit{AfriMMLU} \\
            \midrule
            \textit{Prompt} & \textbf{0.1540} & \textbf{0.0987} & {0.0227} & {0.3356} & {0.1953} & \textbf{0.0698} & {0.0227} & {0.3339} & \textbf{-} & \textbf{0.0672} & {0.0160} & {0.3356} & \textbf{0.2180} \\
            \textit{Translate} & 0.0432 & 0.0165 & \textbf{0.0400} & \textbf{0.3511} & \textbf{0.2253} & 0.0285 & \textbf{0.0467} & \textbf{0.3894} & \textbf{0.2107} & 0.0249 & \textbf{0.0533} & \textbf{0.3894} & 0.1967 \\
            \textit{LAFT} & 0.0778 & 0.0000 & 0.0120 & 0.3294 & 0.1573 & 0.0000 & 0.0000 & 0.0000 & - & 0.0000 & 0.0000 & 0.0000 & 0.0000 \\
            \textit{FOCUS} & 0.0187 & 0.0000 & 0.0080 & 0.1844 & 0.0727 & 0.0000 & 0.0000 & 0.0000 & 0.0000 & 0.0000 & 0.0000 & 0.0000 & 0.0000 \\
            \textit{LAIT} & 0.0159 & 0.0000 & 0.0147 & - & 0.0073 & 0.0000 & 0.0200 & - & - & 0.0034 & 0.0107 & - & 0.0180 \\ \\ \\ \\
        
            \multicolumn{13}{c}{{\emph{\textbf{MPT-7B}}}} \\
            \cmidrule{2-14} \\
            & \multicolumn{5}{c}{\textbf{hau}} & \multicolumn{4}{c}{\textbf{lug}} & \multicolumn{4}{c}{\textbf{kin}} \\
            \cmidrule(lr){2-6} \cmidrule(lr){7-10} \cmidrule(lr){11-14}
            & \textit{XL-Sum} & \textit{MasakhaNER} & \textit{AfriMGSM} & \textit{AfriXNLI} & \textit{AfriMMLU} & \textit{MasakhaNER} & \textit{AfriMGSM} & \textit{AfriXNLI} & \textit{AfriMMLU} & \textit{MasakhaNER} & \textit{AfriMGSM} & \textit{AfriXNLI} & \textit{AfriMMLU} \\
            \midrule
            \textit{Prompt} & \textbf{0.1304} & \textbf{0.1702} & {0.0253} & {0.3356} & {0.1987} & \textbf{0.1820} & {0.0147} & {0.0017} & \textbf{-} & \textbf{0.1820} & \textbf{0.0213} & \textbf{0.2756} & \textbf{0.2200} \\
            \textit{Translate} & 0.0443 & 0.0085 & \textbf{0.0427} & \textbf{0.3417} & 0.1627 & 0.0265 & \textbf{0.0200} & \textbf{0.0783} & \textbf{0.1013} & 0.0168 & 0.0200 & 0.1467 & 0.1627 \\
            \textit{LAFT} & 0.0621 & 0.0000 & 0.0053 & 0.3061 & 0.0627 & 0.0000 & 0.0000 & 0.0000 & - & 0.0000 & 0.0000 & 0.0000 & 0.0000 \\
            \textit{FOCUS} & 0.0138 & 0.0000 & 0.0053 & 0.0150 & \textbf{0.2713} & 0.0000 & 0.0000 & 0.0000 & 0.0000
 & 0.0000 & 0.0000 & 0.0000 & 0.0000
 \\
            \textit{LAIT} & 0.0652 & 0.0000 & 0.0000 & - & 0.0000 & 0.0000 & 0.0000 & - & - & 0.0000 & 0.0013 & 0.0006 & 0.0000 \\ \\ \\ \\
        
            \multicolumn{13}{c}{{\emph{\textbf{Phi-2}}}} \\
            \cmidrule{2-14} \\
            & \multicolumn{5}{c}{\textbf{hau}} & \multicolumn{4}{c}{\textbf{lug}} & \multicolumn{4}{c}{\textbf{kin}} \\
            \cmidrule(lr){2-6} \cmidrule(lr){7-10} \cmidrule(lr){11-14}
            & \textit{XL-Sum} & \textit{MasakhaNER} & \textit{AfriMGSM} & \textit{AfriXNLI} & \textit{AfriMMLU} & \textit{MasakhaNER} & \textit{AfriMGSM} & \textit{AfriXNLI} & \textit{AfriMMLU} & \textit{MasakhaNER} & \textit{AfriMGSM} & \textit{AfriXNLI} & \textit{AfriMMLU} \\
            \midrule
            \textit{Prompt} & \textbf{0.1720} & \textbf{0.1382} & {0.0200} & \textbf{0.3372} & {0.2007} & \textbf{0.0817} & {0.0213} & \textbf{0.3272} & \textbf{-} & \textbf{0.0781} & {0.0213} & \textbf{0.3272} & {0.2053} \\
            \textit{Translate} & 0.0410 & 0.0239 & \textbf{0.1080} & 0.3267 & \textbf{0.2587} & 0.0485 & \textbf{0.0587} & 0.2656 & \textbf{0.2280} & 0.0505 & \textbf{0.1080} & 0.3267 & \textbf{0.2133} \\
            \textit{LAFT} & 0.0930 & 0.0000 & 0.0107 & 0.0511 & 0.1400 & 0.0000 & 0.0000 & 0.0000 & - & 0.0000 & 0.0000 & 0.0000 & 0.0000 \\
            \textit{FOCUS} & 0.0274 & 0.0000 & 0.0067 & 0.1378 & 0.2033 & 0.0000 & 0.0000 & 0.0000 & 0.0000 & 0.0000 & 0.0000 & 0.0000 & 0.0000 \\
            \textit{LAIT} & 0.1079 & 0.0000 & 0.0013 &-  &  0.0020 & 0.0000 & 0.0040 & - & - & 0.0000 & 0.0040 & - & 0.0327 \\
        
            \bottomrule
        \end{tabular}

    \end{table*}

\clearpage
\begin{table*}[]
\scriptsize
\centering
\caption{Downstream task performance in each training setting for Burmese (\texttt{bur}) and Thai (\texttt{tha}) averaged over 3 runs for all models. We report ROUGE-L for XL-Sum, F1-score for WikiANN, and accuracy for MyanmarXNLI, XStoryCloze, MGSM, XNLI, XQUAD, and XCOPA. XL-Sum and WikiANN are language specific.}
\label{tab:fullresults1}

\begin{tabular}{lcccc|cccccc}
        \toprule
        \multicolumn{11}{c}{{\emph{\textbf{LLaMa-2 7B}}}} \\
        \cmidrule{2-11}
        \multirow{2}{*}{\textit{Experiment}} & \multicolumn{4}{c}{\textbf{bur}} & \multicolumn{6}{c}{\textbf{tha}} \\
        \cmidrule(lr){2-5} \cmidrule(lr){6-11}
        & \textit{XL-Sum} & \textit{MyanmarXNLI} & \textit{XStoryCloze} & \textit{WikiANN} & \textit{XL-Sum} & \textit{MGSM} & \textit{XNLI} & \textit{WikiANN} & \textit{XQUAD} & \textit{XCOPA} \\
        \midrule
        \textit{Prompt} & - & - & - & - & - & - & - & - & - & \textbf{0.5233} \\
        \textit{Translate} & \textbf{0.0154} & \textbf{0.1519} & {0.3896} & \textbf{0.0847} & \textbf{0.0296} & \textbf{0.0120} & \textbf{0.2774} & \textbf{0.0053} & \textbf{0.0000} & 0.5013 \\
        \textit{LAFT} & - & - & - & - & - & -  & - & - & - & - \\
        \textit{FOCUS} & 0.0050 & 0.0726 & \textbf{0.1738} & 0.0168 & - & - & - & - & - & - \\
        \textit{LAIT} & - & - & - & - & - & - & -  & - & - & 0.2927 \\ \\ \\ \\
        \multicolumn{10}{c}{{\emph{\textbf{MPT-7B}}}} \\
        \cmidrule{2-10}
        \multirow{2}{*}{\textit{Experiment}} & \multicolumn{3}{c}{\textbf{bur}} & \multicolumn{6}{c}{\textbf{tha}} \\
        \cmidrule(lr){2-5} \cmidrule(lr){6-11}
        & \textit{XL-Sum} & \textit{MyanmarXNLI} & \textit{XStoryCloze} & \textit{WikiANN} & \textit{XL-Sum} & \textit{MGSM} & \textit{XNLI} & \textit{WikiANN} & \textit{XQUAD} & \textit{XCOPA} \\
        \midrule
        \textit{Prompt} & - & - & - & - & - & - & - & - & - & 0.4333 \\
        \textit{Translate} & \textbf{0.0157} & \textbf{0.1419} &  {0.3850} & \textbf{0.0395} & \textbf{0.0246} & \textbf{0.0120} & \textbf{0.2464} & \textbf{0.0059} & \textbf{0.0000} & \textbf{0.5193} \\
        \textit{LAFT} & - & - & - & - & - & - & - & - & - & 0.3827 \\
        \textit{FOCUS} & 0.0047 & - & \textbf{0.4622} & 0.0000 & 0.0087 & 0.0053 & 0.0599 & - & - & 0.3560 \\
        \textit{LAIT} & - & - & - & - & - & - & - & - & - & 0.0307 \\ \\ \\ \\       
        \multicolumn{10}{c}{{\emph{\textbf{Phi-2}}}} \\
        \cmidrule{2-10}
        \multirow{2}{*}{\textit{Experiment}} & \multicolumn{3}{c}{\textbf{bur}} & \multicolumn{6}{c}{\textbf{tha}} \\
        \cmidrule(lr){2-5} \cmidrule(lr){6-11}
        & \textit{XL-Sum} & \textit{MyanmarXNLI} & \textit{XStoryCloze} & \textit{WikiANN} & \textit{XL-Sum} & \textit{MGSM} & \textit{XNLI} & \textit{WikiANN} & \textit{XQUAD} & \textit{XCOPA} \\
        \midrule
        \textit{Prompt} & - & - & - & - & - & - & - & - & - & 0.4473 \\
        \textit{Translate} & \textbf{0.0100} & \textbf{0.1448} & 0.1090 & \textbf{0.0278} & \textbf{0.0270} & \textbf{0.0147} & \textbf{0.2735} & \textbf{0.0067} & \textbf{0.0019} & \textbf{0.4960} \\
        \textit{LAFT} & - & - & - & - & - & - & - & - & - & - \\
        \textit{FOCUS} & 0.0047 & - & \textbf{0.4013} & 0.0000 & 0.0136 & 0.0000 & 0.0464 & - & - & 0.3733 \\
        \textit{LAIT} & - & - & - & - & - & - & - & - & - & - \\
        \bottomrule
    \end{tabular}
    \label{tab:results}
\end{table*}

    \clearpage

\begin{table*}
\centering
\small
\begin{tabular}{p{0.3\columnwidth}p{0.72\columnwidth}p{0.50\columnwidth}p{0.4\columnwidth}}
\toprule
\textbf{Method} & \textbf{Dataset}  & \textbf{Average Input Perplexity} & \textbf{VOR} \\
\midrule
\multirow[t]{6}{*}{Prompt} & AfriMGSM & 6.43e+01 & 0.9450 \\
 & AfriMMLU & 3.59e+01 & 0.9703 \\
 & AfriXNLI & 2.51e+01 & 0.8675 \\
 & XCOPA & 3.05e+00 & 0.9164 \\
 & XL-Sum & 8.39e+01 & 1.0000 \\
 & masakhaNER & 1.61e+01 & 0.6567 \\
\cline{1-4}
\multirow[t]{12}{*}{Translate} & AfriMGSM & 1.77e+01 & 0.8990\\
 & AfriMMLU & 1.17e+01 & 0.8353 \\
 & AfriXNLI & 1.38e+01 & 0.7818 \\
 & MGSM & 1.40e+01 & 0.8382 \\
 & MyanmarXNLI & 1.94e+01 & 0.0000 \\
 & XCOPA & 2.12e+01 & 0.9636 \\
 & XL-Sum & 2.63e+01 & 1.0000 \\
 & XNLI & 1.68e+01 & 0.9487 \\
 & XQUAD & 1.53e+01 & 1.0000 \\
 & XStoryCloze & 1.37e+01 & 0.0000 \\
 & masakhaNER & 1.48e+01 & 0.5415 \\
 & wikiANN & 1.35e+01 & 0.4207 \\
\cline{1-4}
\multirow[t]{6}{*}{LAFT} & AfriMGSM & 6.56e+01 & 0.1684 \\
 & AfriMMLU & 8.08e+01 & 0.2348 \\
 & AfriXNLI & 7.70e+01 & 0.2257 \\
 & XCOPA & 3.66e+00 & 0.8730 \\
 & XL-Sum & 1.19e+01 & 1.0000\\
 & masakhaNER & 2.18e+02 & 0.0211 \\
\cline{1-4}
\multirow[t]{11}{*}{FOCUS} & AfriMGSM & 6.58e+03 & 0.1393 \\
 & AfriMMLU & 5.69e+03 & 0.2526 \\
 & AfriXNLI & 6.48e+03 & 0.1105 \\
 & MGSM & 3.38e+01 & 0.3453 \\
 & MyanmarXNLI & 8.34e+01 & 0.2203 \\
 & XCOPA & 3.07e+01 & 0.6927 \\
 & XL-Sum & 7.53e+01 & 1.0000 \\
 & XNLI & 1.63e+01 & 0.1516 \\
 & XStoryCloze & 5.47e+01 & 0.7326 \\
 & masakhaNER & 4.20e+03 & 0.0375 \\
 & wikiANN & 9.28e+01 & 0.4856 \\
\cline{1-4}
\multirow[t]{6}{*}{LAIT} & AfriMGSM & 1.94e+09 & 0.4010 \\
 & AfriMMLU & 3.15e+10 & 0.0411 \\
 & AfriXNLI & 1.53e+05 & 0.1470 \\
 & XCOPA & 2.55e+06 & 0.3237 \\
 & XL-Sum & 1.93e+10 & 1.0000 \\
 & masakhaNER & 3.19e+07 & 0.0341 \\
\cline{1-4}
\bottomrule
\end{tabular}
\caption{Average Input Perplexity and VOR Scores}
\end{table*}

\begin{table*}[htbp]
\centering
\small
\begin{tabular}{p{0.2\columnwidth}p{1.8\columnwidth}}
\toprule
\textbf{Lang.}  & \textbf{Example Input + \textcolor{blue}{Output}} \\ \midrule
\texttt{hau} &  Fitar da amsar karshe kawai ga tambayar lissafi. 
 Leah nada 32 chaculet, yar uwarta kuma 42.gudanawa suka rage musu? --> 39

Fitar da amsar karshe kawai ga tambayar lissafi. 
 Agwagin Janet suna yin kwai 16 a kullun. Tana yin karin kumallo da guda uku kowace safiya, sannan tana gasawa kawayenta guda hudu kullum. A kullum takan sayar da ragowar a kasuwar manoma akan dala 2 akan kowane kwai. Dala nawa take samu a kullum a kasuwar manoma? --> \textcolor{blue}{\textbf{29}} \\ \hline
\texttt{kin} &  Ibisohoka gusa igisubizo cyanyuma kubibazo byimibare. 
 Leah afite shokola 32 naho umuvandimwe we afite 42. Nibarya 35 bazaba basigaranye shokola zingahe zose hamwe? --> 39

Ibisohoka gusa igisubizo cyanyuma kubibazo byimibare. 
 Igishuhe cya Jane gitera amajyi 16 ku munsi,  buri mugitondo aryamo atatu kandi akora  umugati winshutiye akoresheje ane, agurisha asigaye mwisoko ryabahinzi buri munsi kugichiro cya 2 kuri buri jyi. Na ngahe mumadolali yinjiza ku munsi  mwisoko ryabahinzi ? --> \textcolor{blue}{\textbf{39}} \\ \hline
\texttt{lug} &  Fulumya eky’okuddamu ekisembayo kyokka ku kibuuzo ky’okubala. 
 Leah yalina kyokuleeti 32 ate nga muganda we ye yalina 42. Bwe baba nga baalyako 35, baasigazaawo kyokuleeti mmeka bombi omugatte? --> 39

Fulumya eky’okuddamu ekisembayo kyokka ku kibuuzo ky’okubala. 
 Embaata za Janet zibiika amagi 16 buli lunaku. Alya amagi asatu buli lunaku ku kyenkya n'afumbisa amalala ana g'ateeka mu bukkeeki bwa muffin bw'akolera mikwano gye. Agasigadde agatunda mu katale k'abalimi n'abalunzi buli lunaku nga buli ggi alitunda \$2. Afuna ssente mmeka buli lunaku mu katale k'abalimi n'abalunzi? --> \textcolor{blue}{\textbf{19}} \\ \hline
 \bottomrule
\end{tabular}
\caption{Example few-shot prompts and their respective model outputs for the Prompt adaptation method on AfriMGSM. We use the same prompts for all models, but the reported outputs here are from one of the random seeds in the LLaMa2-7B experiments.}
\end{table*}

\begin{table*}[htbp]
\centering
\small
\begin{tabular}{p{0.2\columnwidth}p{1.8\columnwidth}}
\toprule
\textbf{Lang.}  & \textbf{Example Input + \textcolor{blue}{Output}} \\ \midrule
\texttt{hau} &  Zabi zabin amsa daidai: A, B, C, ko D. 
 Wani masanin kimiyya ya auna dayamita na gashin mutum hudu. Dayamitocin, a ma'anin milimita, sune 0.091, 0.169, 0.17, da 0.023. Wanne in'ikwaliti ne ya kwatanta biyu daga dayamitocin biyu na gashin dan adam? 
 A: 0.17 > 0.023 
 B: 0.091 < 0.023 
 C: 0.169 > 0.17 
 D: 0.17 < 0.091 --> A

Zabi zabin amsa daidai: A, B, C, ko D. 
 Menene matsayin p a cikin 24 = 2p? 
 A: p = 4 
 B: p = 8 
 C: p = 12 
 D: p = 24 --> \textcolor{blue}{\textbf{A}} \\ \hline
\texttt{kin} &  Tora igisubizo gikwiye: A, B, C, cyangwa D. 
 Umuhanga yapimye diameter yimisatsi ine yabantu. Diameter, muri milimetero, yari 0.091, 0.169, 0.17, na 0.023. Ni ubuhe busumbane bugereranya neza diameter yimisatsi ibiri muriyo misatsi yabantu? 
 A: 0.17 > 0.023 
 B: 0.091 < 0.023 
 C: 0.169 > 0.17 
 D: 0.169 > 0.17 --> A

Tora igisubizo gikwiye: A, B, C, cyangwa D. 
 Nakahe gaciro ka p muri 24 = 2p?  
 A: p = 4 
 B: p = 8 
 C: p = 12 
 D: p = 24 --> \textcolor{blue}{\textbf{A}} \\ \hline
 \bottomrule
\end{tabular}
\caption{Example few-shot prompts and their respective model outputs for the Prompt adaptation method on AfriMMLU. We use the same prompts for all models, but the reported outputs here are from one of the random seeds in the LLaMa2-7B experiments.}
\end{table*}

\begin{table*}[htbp]
\centering
\small
\begin{tabular}{p{0.2\columnwidth}p{1.8\columnwidth}}
\toprule
\textbf{Lang.}  & \textbf{Example Input + \textcolor{blue}{Output}} \\ \midrule
\texttt{hau} &  kayyade idan hasashen ya bi jigo a hankali. Fitowa 0 don entailment, 1 don tsaka tsaki, ko 2 don sabani. 
 A Tsakanin 1936 da 1940 Greece na karkashin mulkin kama karya na loannus Metaxas, Ana iya tunawa da sutin (a'a)  da ya amsa dashi zuwa ga Mussolini ultimatum yayi mubaya'a a 1940. Tattalin arzikin Greece bai yi kyau ba a karkashin mulkin kama karya na soja na Metaxas. --> 1

kayyade idan hasashen ya bi jigo a hankali. Fitowa 0 don entailment, 1 don tsaka tsaki, ko 2 don sabani. 
 Wadannan rikirkitattun na'urar kwayoyin halitta sun samo asali ne saboda zabin su a haka zai iya canza yanayi su gaba daya dan haka su kwayoyin halitta suna taruwa lokacin da yanayin su na gaba daya ya habaka kuma ya canza da yanayi da suke. Duk na'urorin kwayoyin halitta suna da wahalar sha'ani. --> 2

kayyade idan hasashen ya bi jigo a hankali. Fitowa 0 don entailment, 1 don tsaka tsaki, ko 2 don sabani. 
 masu son karatu musamman wanda suka manne a ajin karatun sanin tattalin arziki da na na'ura mai kwakwalwa basu da wani alfanu nan gaba. masu san karatu basu da wani alfanu. --> 0

kayyade idan hasashen ya bi jigo a hankali. Fitowa 0 don entailment, 1 don tsaka tsaki, ko 2 don sabani. 
 Gaskiya bana ba tunanin sa amma na fusata sosai, kuma dai daga karshe na bige da kara yi masa magana. Ban kara masa magana ba. --> \textcolor{blue}{\textbf{1}} \\ \hline
\texttt{kin} &  Menya niba hypothesis ikurikiza ishingiro. Ibisohoka 0 kubisobanuro, 1 kubutabogamye, cyangwa 2 kubivuguruza. 
 Hagati ya 1936 na 1940 Ubugereki bwari ku butegetsi bw'igitugu bwa gisirikare bwa Ioannis Metaxas, bwibukwa kubera echi yumvikana (oya ) yatanze asubiza ultimatum ya Mussolini yokwiyegurira mu 1940. Ubukungu bw'Ubugereki ntabwo bwaribumeze neza kubutegetsi bwigitugu bwa gisirikare bwa Metaxas --> 1

Menya niba hypothesis ikurikiza ishingiro. Ibisohoka 0 kubisobanuro, 1 kubutabogamye, cyangwa 2 kubivuguruza. 
 Izi nzego zo murwego rwohejuru rwibikoresho bya molekile bivuka kubera ko gutoranya bisanzwe gushobora gukora kumitungo rusange yibintu bya molekile iyo iyo mitungo rusange yongerewe imbaraga zo guhuza n'imihindagurikire y'ikirere.
 Ibikoresho byose bya molekile biba bgoranye --> 2

Menya niba hypothesis ikurikiza ishingiro. Ibisohoka 0 kubisobanuro, 1 kubutabogamye, cyangwa 2 kubivuguruza. 
 Aba hanga bahatamye cyane mubyubukungu n'imyijyire ya kopyuta ,ninabo bafite ukwizera gucye Aba hanga bakompyuta ntakizere bafite --> 0

Menya niba hypothesis ikurikiza ishingiro. Ibisohoka 0 kubisobanuro, 1 kubutabogamye, cyangwa 2 kubivuguruza. 
 Urebye, ntabwo nigeze ntekereza kuribyo, ariko narumiwe cyane, ndangije nongeye kumuvugisha tena Ntabwo narinongera kumuvugisha  --> \textcolor{blue}{\textbf{1}} \\ \hline
\texttt{lug} &  Salawo oba endowooza (hypothesis) egoberera mu ngeri entegeerekeka (logically) ensonga (premise). Ekifulumizibwa 0 ku entailment, 1 ku neutral, oba 2 ku contradiction. 
 Mu mutendera oguddako wansi, ddayirekita w'akabinja ka al Qaeda mu kitongole kya CIA mu kiseera ekyo yajjukira nti yali talowooza nti gwali mulimu gwe okulagira ekirina okukolebwa oba obutakolebwa.   Ddayirekita w'ekitundu ekyo yali tayagala kwenyigira mu kuddukanya ekyo ekyali kikolebwa. --> 0

Salawo oba endowooza (hypothesis) egoberera mu ngeri entegeerekeka (logically) ensonga (premise). Ekifulumizibwa 0 ku entailment, 1 ku neutral, oba 2 ku contradiction. 
 Mary Traill ajja kukikugambako. Nkimanyiiko. --> 2

Salawo oba endowooza (hypothesis) egoberera mu ngeri entegeerekeka (logically) ensonga (premise). Ekifulumizibwa 0 ku entailment, 1 ku neutral, oba 2 ku contradiction. 
 Naye nedda, omanyi sikaati na bulawuzi oba ng'olaba kiteeteeyi wano, naye kirungi gyendi okukolera awaka kubanga mba nsobola n'okwambala engoye z'omunda. Ssambala kintu kirala kyonna okuggyako essweta bwe nkolera ewaka. --> 1

Salawo oba endowooza (hypothesis) egoberera mu ngeri entegeerekeka (logically) ensonga (premise). Ekifulumizibwa 0 ku entailment, 1 ku neutral, oba 2 ku contradiction. 
 Kale nno, ekyo si na kye nnabadde ndowoozaako, naye olw'okuba nnabadde mu mbeera ey'okusoberwa, nnawunzise nzizeemu okwogera naye. Sinnaddamu kwogerako naye. --> \textcolor{blue}{\textbf{0}} \\ \hline
 \bottomrule
\end{tabular}
\caption{Example few-shot prompts and their respective model outputs for the Prompt adaptation method on AfriXNLI. We use the same prompts for all models, but the reported outputs here are from one of the random seeds in the LLaMa2-7B experiments.}
\end{table*}

\clearpage            % flush all pending floats
\includepdf[pages=1]{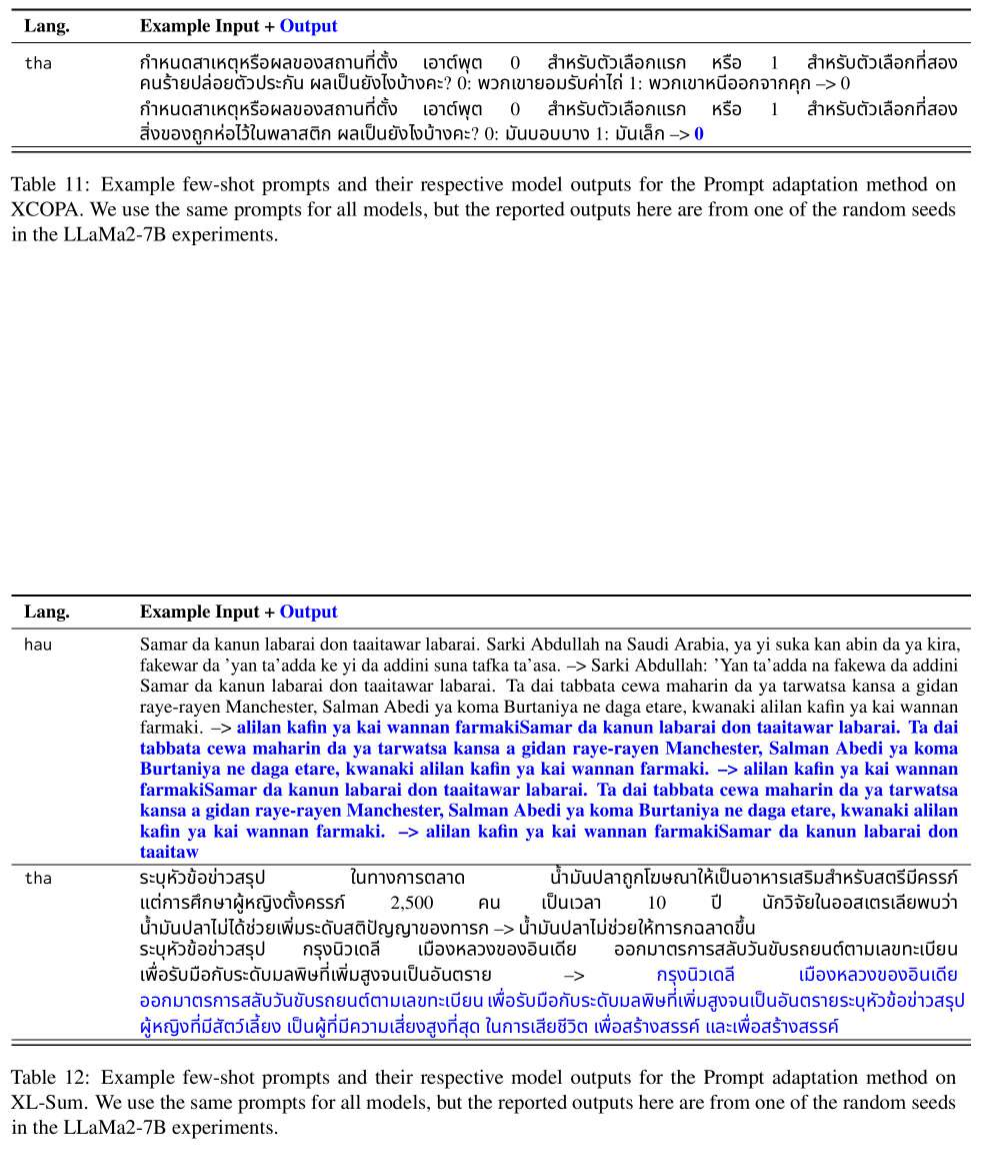}
\clearpage            % isolate this page from others

\setcounter{table}{12}
\begin{table*}[htbp]
\centering
\small
\begin{tabular}{p{0.2\columnwidth}p{1.8\columnwidth}}
\toprule
\textbf{Lang.}  & \textbf{Example Input + \textcolor{blue}{Output}} \\ \midrule
\texttt{hau} &  Sanya kowace kalma a cikin jumla mai zuwa tare da alamar NER. 
 Sai dai mai sharhi akan harkokin siyasa na kasar Delphin Kapaya ya ce yadda kotun ta gudanar da wannan shara’ar shine zai nuna irin mataki nagaba da magoya bayan Beman zasu dauka . --> (Sai,O), (dai,O), (mai,O), (sharhi,O), (akan,O), (harkokin,O), (siyasa,O), (na,O), (kasar,O), (Delphin,B-PER), (Kapaya,I-PER), (ya,O), (ce,O), (yadda,O), (kotun,O), (ta,O), (gudanar,O), (da,O), (wannan,O), (shara’ar,O), (shine,O), (zai,O), (nuna,O), (irin,O), (mataki,O), (nagaba,O), (da,O), (magoya,O), (bayan,O), (Beman,B-PER), (zasu,O), (dauka,O), (.,O)

Sanya kowace kalma a cikin jumla mai zuwa tare da alamar NER. 
 Ya kuma yaba da shawarar da bangaren al - Barnawi na Boko Haram ya yanke na sassautawa a gwagwarmayarsu . --> \textcolor{blue}{\textbf{[]}} \\ \hline
\texttt{kin} &  Shyira buri jambo mu nteruro ikurikira hamwe na tagi ya NER. 
 Amazon iteganya gushora miliyari 6 , 5 z’amadorali mu bikorwa byo gucururiza kuri internet , ndetse ngo ikaba izoroherwa no gukoresha internet ya Bharti ku giciro gito mu gihe ibigo byombi bizaba byemeranyije amasezerano byifuza . --> (Amazon,B-ORG), (iteganya,O), (gushora,O), (miliyari,O), (6,O), (,,O), (5,O), (z’amadorali,O), (mu,O), (bikorwa,O), (byo,O), (gucururiza,O), (kuri,O), (internet,O), (,,O), (ndetse,O), (ngo,O), (ikaba,O), (izoroherwa,O), (no,O), (gukoresha,O), (internet,O), (ya,O), (Bharti,B-ORG), (ku,O), (giciro,O), (gito,O), (mu,O), (gihe,O), (ibigo,O), (byombi,O), (bizaba,O), (byemeranyije,O), (amasezerano,O), (byifuza,O), (.,O)

Shyira buri jambo mu nteruro ikurikira hamwe na tagi ya NER. 
 Bazwi mu cyo bise ‘Morning Worship’ aho baririmba ibihangano bitandukanye byo mu gitabo bigafasha benshi . --> \textcolor{blue}{\textbf{['(Morning,ORG)', '(worship,ORG)', '(mu,ORG)', '(cyo,ORG)', '(bise,ORG)', '(aho,ORG)', '(baririmba,ORG)', '(ibihangano,ORG)', '(bitandukanye,ORG)', '(byo,ORG)', '(gitabo,ORG)', '(bigafasha,ORG)', '(benshi,ORG)', '(.,ORG)', '(Morning,ORG)', '(worship,ORG)', '(mu,ORG)', '(cyo,ORG)', '(bise,ORG)']}} \\ \hline
\texttt{lug} &  Buli kigambo mu sentensi eno wammanga giteekeko akabonero kaakyo aka NER. 
 Abantu abaatuwa obuyambi bampa sikaala okugenda mu Amerika okusoma diguli eyookubiri olwo bizinensi yenkoko ne ngiwa mukwano gwange Geoffrey Lwanga nga kati mu kiseera kino alina enkoko ezisoba mu 7000 ezamagi . --> (Abantu,O), (abaatuwa,O), (obuyambi,O), (bampa,O), (sikaala,O), (okugenda,O), (mu,O), (Amerika,B-LOC), (okusoma,O), (diguli,O), (eyookubiri,O), (olwo,O), (bizinensi,O), (yenkoko,O), (ne,O), (ngiwa,O), (mukwano,O), (gwange,O), (Geoffrey,B-PER), (Lwanga,I-PER), (nga,O), (kati,O), (mu,O), (kiseera,B-DATE), (kino,I-DATE), (alina,O), (enkoko,O), (ezisoba,O), (mu,O), (7000,O), (ezamagi,O), (.,O)

Buli kigambo mu sentensi eno wammanga giteekeko akabonero kaakyo aka NER. 
 Ono ye waffe era kampeyini ze okuziyimirizaawo tujja kwesondamu ensimbi ezinamuyamba okukuba ebipande ebipande nokukola emirimu emirara , Rose Namuli akolera ku katale ka Pepsi oluvanyuma namuwa 2 , 000 . --> \textcolor{blue}{\textbf{['( Rose Namuli, PER PER)', '(Pepsi,LOC)']}} \\ \hline
 \bottomrule
\end{tabular}
\caption{Example few-shot prompts and their respective model outputs for the Prompt adaptation method on masakhaNER. We use the same prompts for all models, but the reported outputs here are from one of the random seeds in the LLaMa2-7B experiments.}
\end{table*}

\begin{table*}[htbp]
\centering
\small
\begin{tabular}{p{0.2\columnwidth}p{1.8\columnwidth}}
\toprule
\textbf{Lang.}  & \textbf{Example Input + \textcolor{blue}{Output}} \\ \midrule
\texttt{hau} &  Just give the final answer to the math question. Leah chose 32 chocolates, and her sister 42. Do they have any left? --> 39

Just to give the final answer to the math question. Janet's aunt makes 16 eggs a day. She makes three for breakfast each morning, and then she brews four for her friends every day. She sells the leftovers at the farmers market for \$2 per egg. How many dollars does she make a day at the farmers market? --> \textcolor{blue}{\textbf{16}} \\ \hline
\texttt{kin} &  Only the last answer to the math questions is given. Leah has 32 chocolates and her brother has 42. If there are at least 35 chocolates, how many chocolates will they have left in total? --> 39

It only comes out the last answer to the math questions. Jane's salary is 16 cents a day, she sleeps in three and makes a nice loaf of bread with four, she sells the rest at the farmers market every day for 2 cents a day. How many dollars does she make a day at the farmers market? --> \textcolor{blue}{\textbf{18}} \\ \hline
\texttt{lug} &  Write out the last answer to the number question. Leah had 32 and her brother had 42. If they ate 35, how many were left? --> 39

Give only the last answer to the math question. Janet's chickens lay 16 eggs a day. She eats three eggs a day for breakfast and cooks four more for her friends' muffin. She sells them at the farmers market every day for \$2 apiece. How much money does she make a day at the farmers market? --> \textcolor{blue}{\textbf{12}} \\ \hline
 \bottomrule
\end{tabular}
\caption{Example few-shot prompts and their respective model outputs for the Translate adaptation method on AfriMGSM. We use the same prompts for all models, but the reported outputs here are from one of the random seeds in the LLaMa2-7B experiments.}
\end{table*}

\begin{table*}[htbp]
\centering
\small
\begin{tabular}{p{0.2\columnwidth}p{1.8\columnwidth}}
\toprule
\textbf{Lang.}  & \textbf{Example Input + \textcolor{blue}{Output}} \\ \midrule
\texttt{hau} &  Choose the correct answer: A, B, C, or D. A scientist measured the diameter of four human hairs. The diameters, in millimeters, are 0.091, 0.169, 0.17, and 0.023. -->  A

Choose the correct answer option: A, B, C, or D. What is the position of p in 24 = 2p? A: p = 4 B: p = 8 C: p = 12 D: p = 24 --> \textcolor{blue}{\textbf{A}} \\ \hline
\texttt{kin} &  The diameter, in millimeters, was 0.091, 0.169, 0.17, and 0.023. What inequality best represents the diameter of two of the hairs in a human hair? A: 0.17 > 0.023 B: 0.091 < 0.023 C: 0.169 > 0.17 D: 0.169 > 0.17 --> A

Find the correct answer: A, B, C, or D. What is the value of p in 24 = 2p? A: p = 4 B: p = 8 C: p = 12 D: p = 24 --> \textcolor{blue}{\textbf{A}} \\ \hline
\texttt{lug} &  Choose the correct answer: A, B, C, or D. The scientist measured the width of four human hair strands. The fractional lengths are 0.091, 0.169, 0.17, and 0.023. What is the relationship between the exact values that can be used to compare the width of two human hair strands? A: 0.17 > 0.023 B: 0.091 < 0.023 C: 0.169 > 0.17 D: 0.17 < 0.091 --> A

Choose the correct answer: A, B, C, or D. What is the value of p in 24 = 2p? A: p = 4 B: p = 8 C: p = 12 D: p = 24 --> \textcolor{blue}{\textbf{A}} \\ \hline
 \bottomrule
\end{tabular}
\caption{Example few-shot prompts and their respective model outputs for the Translate adaptation method on AfriMMLU. We use the same prompts for all models, but the reported outputs here are from one of the random seeds in the LLaMa2-7B experiments.}
\end{table*}

\begin{table*}[htbp]
\centering
\small
\begin{tabular}{p{0.2\columnwidth}p{1.8\columnwidth}}
\toprule
\textbf{Lang.}  & \textbf{Example Input + \textcolor{blue}{Output}} \\ \midrule
\texttt{hau} &  Determine if the prediction follows the theme closely. Output 0 for entailment, 1 for neutrality, or 2 for contradiction. Between 1936 and 1940 Greece was under the dictatorship of loannus Metaxas, It may be remembered for the sutin (no) he responded to Mussolini's ultimatum to capitulate in 1940. The Greek economy did not fare well under Metaxas' military dictatorship. --> 1

Determine if the prediction follows the theme closely. Exit 0 for entailment, 1 for neutrality, or 2 for contradiction. These complex molecular machines evolved because their selection could change their overall state so that their molecular assemblies when their overall state evolved and changed with the state they were in. All molecular machines are complex. --> 2

Determine if the prediction follows the theme closely. Output 0 for entailment, 1 for neutrality, or 2 for contradiction. amateur readers especially who stuck to the economics and computer science classes had no future advantages. knowledgeable readers had no advantages. --> 0

Determine if the prediction follows the topic carefully. Output 0 for entailment, 1 for neutrality, or 2 for contradiction. I honestly didn't think about it but I was very angry, and I finally snapped and spoke to him again. I didn't speak to him again. --> \textcolor{blue}{\textbf{1}} \\ \hline
\texttt{kin} &  Determine whether the hypothesis is supported. Outputs 0 for explanation, 1 for neutrality, or 2 for contradiction. Between 1936 and 1940 Greece was under the military dictatorship of Ioannis Metaxas, remembered for his eloquent (no) response to Mussolini's ultimatum to surrender in 1940. The Greek economy did not adapt well to Metaxas' military dictatorship --> 1

Determine whether the hypothesis is supported. Outputs 0 for explanation, 1 for neutrality, or 2 for contradiction. These higher-order classes of molecular properties arise because natural selection can act on the shared properties of molecular entities when those shared properties are enhanced to accommodate the changing environment. All molecular entities are complex --> 2

Determine whether the hypothesis is supported. Outputs 0 for explanation, 1 for neutrality, or 2 for contradiction. These nations are highly fragmented economically and computer-centrically, or have little faith in the computer-centricity of their nations. These nations have no confidence in the computer-centricity of their nations --> 0

Determine whether the hypothesis is supported. Outputs 0 for explanation, 1 for objection, or 2 for contradiction. Actually, I never thought about that, but I was very surprised, so I talked to him again. I never spoke to him again --> \textcolor{blue}{\textbf{0}} \\ \hline
\texttt{lug} &  Determine whether a hypothesis follows a premise logically. The output is 0 for entailment, 1 for neutral, or 2 for contradiction. Between 1936 and 1940 Greece was under the dictatorship of Ioannis Metaxas, best remembered for his 'No' response to Mussolini's offer of a hanging sentence after his defeat in 1940. The Greek economy did not fare well during the period under the dictatorship of Metaxas. --> 1

Determine whether a hypothesis follows logically from a premise. The output is 0 for entailment, 1 for neutral, or 2 for contradiction. Higher-order functions arise because the universe has the capacity to do so when it is adapted to do so. All functions are higher-order. --> 2

Determine whether the hypothesis follows logically from the premise. The output is 0 for entailment, 1 for neutral, or 2 for contradiction. People who lack social skills seek safety in economics or computer science classes, and are more likely to live in a hopeless situation. People who lack social skills are hopeless. --> 0

Decide whether the hypothesis follows logically from the premise. The output is 0 for the entailment, 1 for the neutral, or 2 for the contradiction. Well, that's not what I was thinking, but because I was in a state of confusion, I ended up talking to him again. I never spoke to him again. --> \textcolor{blue}{\textbf{1}} \\ \hline
 \bottomrule
\end{tabular}
\caption{Example few-shot prompts and their respective model outputs for the Translate adaptation method on AfriXNLI. We use the same prompts for all models, but the reported outputs here are from one of the random seeds in the LLaMa2-7B experiments.}
\end{table*}

\begin{table*}[htbp]
\centering
\small
\begin{tabular}{p{0.2\columnwidth}p{1.8\columnwidth}}
\toprule
\textbf{Lang.}  & \textbf{Example Input + \textcolor{blue}{Output}} \\ \midrule
\texttt{tha} &  First, think of a step-by-step way to answer a math problem, then print out the final answer. Challenge: Leah has 32 chocolates and her sister has 42. If both of them ate 35 chocolates, how many chocolates would be left? --> 39

First, think of a step-by-step way to answer a math question, then print out the final answer. Janet's egg lays 16 pounds of eggs a day, she eats three eggs for breakfast every day, and she makes four for her friends every day, she sells the rest at the farmers market every day for \$2 for a fresh egg, how much money does she make from the farmers market per day? --> \textcolor{blue}{\textbf{<NAN>}} \\ \hline
 \bottomrule
\end{tabular}
\caption{Example few-shot prompts and their respective model outputs for the Translate adaptation method on MGSM. We use the same prompts for all models, but the reported outputs here are from one of the random seeds in the LLaMa2-7B experiments.}
\end{table*}

\begin{table*}[htbp]
\centering
\small
\begin{tabular}{p{0.2\columnwidth}p{1.8\columnwidth}}
\toprule
\textbf{Lang.}  & \textbf{Example Input + \textcolor{blue}{Output}} \\ \midrule
\texttt{bur} &  Determine whether the hypothesis is capital compatible: yield 0 for correlation, 1 for neutrality, or 2 for inverse. McKim not only lost due to his many frustrations, but finished third behind Howard \& Cowell. McKim was satisfied that he had finished first --> 2

We can then determine whether the concept is capital compatible, yielding 0 for coherence, 1 for neutrality, or 2 for inversion. We can be surprised that others use language in a simple way, and that it ends on our analytical side and begins on our emotional side.

And I think the really interesting thing is, what can we do about this? I mean, we have to change the people who are going to represent us. And it's so boring, and we know that it's not worth changing our representation, so we shouldn't even try to change it.

And then we have the inverse of the equation, which is the inverse of the equation, and we have the inverse of the equation, which is the inverse of the equation, and we have the inverse of the equation, which is the inverse of the equation, and we have the inverse of the equation, which is the inverse of the equation. --> \textcolor{blue}{\textbf{<NAN>}} \\ \hline
 \bottomrule
\end{tabular}
\caption{Example few-shot prompts and their respective model outputs for the Translate adaptation method on MyanmarXNLI. We use the same prompts for all models, but the reported outputs here are from one of the random seeds in the LLaMa2-7B experiments.}
\end{table*}

\begin{table*}[htbp]
\centering
\small
\begin{tabular}{p{0.2\columnwidth}p{1.8\columnwidth}}
\toprule
\textbf{Lang.}  & \textbf{Example Input + \textcolor{blue}{Output}} \\ \midrule
\texttt{tha} &  Determine the cause and effect of the location, put 0 for the first option or 1 for the second option, the perpetrator releases the hostage, what is the result? 0: They accept the ransom: 1: They escape from prison --> 0

Determine the cause and effect of the location, put 0 for the first option, or 1 for the second option, the object is wrapped in plastic, what is the result? 0: it's thin 1: it's small --> \textcolor{blue}{\textbf{0}} \\ \hline
 \bottomrule
\end{tabular}
\caption{Example few-shot prompts and their respective model outputs for the Translate adaptation method on XCOPA. We use the same prompts for all models, but the reported outputs here are from one of the random seeds in the LLaMa2-7B experiments.}
\end{table*}

\begin{table*}[htbp]
\centering
\small
\begin{tabular}{p{0.2\columnwidth}p{1.8\columnwidth}}
\toprule
\textbf{Lang.}  & \textbf{Example Input + \textcolor{blue}{Output}} \\ \midrule
\texttt{bur} &  Give a headline for your news summary: Myanmar government forces and the KLA forces have been engaged in close combat since the morning of July 7th around the old exit road of Kokraj.

Give a headline for this summary: Japanese government cancels plans to build the stadium that will host the 2020 Tokyo Olympics. --> \textcolor{blue}{\textbf{Give a headline for this summary: The United States and China have agreed to resume trade talks.Give a headline for this summary: The United States and China have agreed to resume trade talks.Give a headline for this summary: The United States and China have agreed to resume trade talks.Give a headline for this summary: The United States and China have agreed to resume trade talks.Give a headline for this summary: The United States and China have agreed to resume trade talks.Give a headline for this summary: The United States and China have agreed to resume trade talks.Give a headline for this summary: The United States and China have agreed to resume trade talks.Give a headline for this summary: The United States and China have agreed to resume trade talks.Give a headline for this summary: The United States and China have agreed to resume trade talks.Give a headline for this summary: The United States and China have agreed to resume trade talks.Give a head}} \\ \hline
\texttt{hau} &  King Abdullah of Saudi Arabia, has criticized what he calls, the religious cover-up of terrorism in his country.

She confirmed that the Manchester nightclub suicide bomber, Salman Abedi, had returned to the UK from overseas, just days before the attack. --> \textcolor{blue}{\textbf{\#\#\# 2017* 2017–18 Saudi-led intervention in Yemen* 2017–18 Saudi-led intervention in Yemen* 2017–18 Saudi-led intervention in Yemen* 2017–18 Saudi-led intervention in Yemen* 2017–18 Saudi-led intervention in Yemen* 2017–18 Saudi-led intervention in Yemen* 2017–18 Saudi-led intervention in Yemen* 2017–18 Saudi-led intervention in Yemen* 2017–18 Saudi-led intervention in Yemen* 2017–18 Saudi-led intervention in Yemen* 2017–18 Saudi-led intervention in Yemen* 2017–18 Saudi-led intervention in Yemen* 2017–18 Saudi-led intervention in Y}} \\ \hline
\texttt{tha} &  In the commercials, fish oil is promoted as a supplement for pregnant women, but after studying 2,500 pregnant women for 10 years, researchers in Australia found that fish oil did not increase the intelligence of the baby -- fish oil did not make the baby smarter.

In a headline summary, India's capital New Delhi has instituted a shift in driving days to license plate numbers to deal with dangerously high pollution levels. --> \textcolor{blue}{\textbf{In a headline summary, a new study finds that the number of people who have died from the coronavirus in the United States is 10 times higher than the official number.In a headline summary, a new study finds that the number of people who have died from the coronavirus in the United States is 10 times higher than the official number.In a headline summary, a new study finds that the number of people who have died from the coronavirus in the United States is 10 times higher than the official number.In a headline summary, a new study finds that the number of people who have died from the coronavirus in the United States is 10 times higher than the official number.In a headline summary, a new study finds that the number of people who have died from the coronavirus in the United States is 10 times higher than the official number.In a headline summary, a new study finds that the number of people who have died from the coronavirus in the United States is 10 times higher than the official number.In a headline summary, a new study finds that the number of}} \\ \hline
 \bottomrule
\end{tabular}
\caption{Example few-shot prompts and their respective model outputs for the Translate adaptation method on XL-Sum. We use the same prompts for all models, but the reported outputs here are from one of the random seeds in the LLaMa2-7B experiments.}
\end{table*}

\begin{table*}[htbp]
\centering
\small
\begin{tabular}{p{0.2\columnwidth}p{1.8\columnwidth}}
\toprule
\textbf{Lang.}  & \textbf{Example Input + \textcolor{blue}{Output}} \\ \midrule
\texttt{tha} &  Considering whether the assumption is logical, putting 0 for participation, 1 for neutrality or 2 for conflict, McKim was very disappointed that he not only lost but came in third behind Howard \& Cauldwell. McKim was delighted because he finished first --> 2

Consider whether the assumption is rational, give it a 0 for participation, a 1 for neutrality or a 2 for conflict, others will still be just amazed at the language and wonder just where our analytical side ends and our emotional side begins.

Consider whether the assumption is rational, put 0 for participation, 1 for neutrality or 2 for conflict, and that's what I think it would be really interesting is what we do about it. I mean, we have to change who represents us. I just know it's boring and not worth changing who represents us, so we shouldn't try to change -- 2

Consider whether the assumption is rational, put 0 for participation, 1 for neutrality or 2 for conflict. Well, I didn't think anything of it, but I was disappointed, and, I went back to talk to him, and I didn't talk to him again. --> \textcolor{blue}{\textbf{<NAN>}} \\ \hline
 \bottomrule
\end{tabular}
\caption{Example few-shot prompts and their respective model outputs for the Translate adaptation method on XNLI. We use the same prompts for all models, but the reported outputs here are from one of the random seeds in the LLaMa2-7B experiments.}
\end{table*}

\begin{table*}[htbp]
\centering
\small
\begin{tabular}{p{0.2\columnwidth}p{1.8\columnwidth}}
\toprule
\textbf{Lang.}  & \textbf{Example Input + \textcolor{blue}{Output}} \\ \midrule
\texttt{tha} &  Considering the context, the mechanisms used to evade the adaptive immune system are complex, the simplest method being to rapidly change the non-essential antigen (amino acids and/or sugars) on the surface of the virus while protecting the important antigen. This method is called antigenic mutation, for example, HIV, which rapidly changes the shape of proteins on its viral coat that are important for entering the target cells of the host. These frequent changes in the antigen can be explained as a failure of the genes that are involved in what the virus is meant to do. This virus has been brought to the ultimate stage of uncontrolled cell detection. The same mechanism can be used to prevent changes in the immune system from itself by preventing the immune system from recognizing other cells that are not immune.

In some cases, it has been proposed to hold a plea bargain for the perpetrators of rape, as in the case of Camden 28, where the accused was offered the opportunity to confess to a crime in order to avoid imprisonment. In some mass arrest situations, activists decided to use the same unity strategy so that everyone could confess with the same plea bargain, but some activists chose to confess to the crime, admitted without any plea bargain, Mahatma Gandhi confessed, and told the court, "I am here... willing to accept the maximum punishment that can be imposed on me for what I consider to be a legal crime, which was planned in advance, but which I consider to be the highest duty of the citizenry to impose on the perpetrator". --> \textcolor{blue}{\textbf{<NAN>}} \\ \hline
 \bottomrule
\end{tabular}
\caption{Example few-shot prompts and their respective model outputs for the Translate adaptation method on XQUAD. We use the same prompts for all models, but the reported outputs here are from one of the random seeds in the LLaMa2-7B experiments.}
\end{table*}

\begin{table*}[htbp]
\centering
\small
\begin{tabular}{p{0.2\columnwidth}p{1.8\columnwidth}}
\toprule
\textbf{Lang.}  & \textbf{Example Input + \textcolor{blue}{Output}} \\ \midrule
\texttt{bur} &  Given the context, choose the best ending for the story: 1 or 2. This Sunday, there is a lot for Amber to do. She has made a list of places to go. She is in a hurry to get ready. She is worried about the time. 1: Amber enjoys the comfortable two-hour breakfast and lunch combination. 2: Amber left the list at home and had to work in a hurry.

Given the context, choose the best ending for the story: 1 or 2. I became a fan of Law and Order in 2011. I had recovered from a stroke. When I got home, I tried to watch every episode. --> \textcolor{blue}{\textbf{<NAN>}} \\ \hline
 \bottomrule
\end{tabular}
\caption{Example few-shot prompts and their respective model outputs for the Translate adaptation method on XStoryCloze. We use the same prompts for all models, but the reported outputs here are from one of the random seeds in the LLaMa2-7B experiments.}
\end{table*}

\begin{table*}[htbp]
\centering
\small
\begin{tabular}{p{0.2\columnwidth}p{1.8\columnwidth}}
\toprule
\textbf{Lang.}  & \textbf{Example Input + \textcolor{blue}{Output}} \\ \midrule
\texttt{hau} &  Place each word in the following sentence with the NER symbol. However, the commentator on national politics Delphin Kapaya says that the way the court handled this ruling will indicate what action the Beman supporters will take. --> (Sai,O), (dai,O), (mai,O), (comment,O), (akan,O), (politik,O), (politik,O), (de,O), (negara,O), (Delphin,B-PER), (Kapaya,I-PER), (ya,O), (ce,O), (how,O), (kotun,O), (ta,O), (gover,O), (da,O), (wannan,O), (shara,O), (Oshine), (O), (Ouna,O), (O), (mat,O), (O), (magoda,O), (Bira,

Place each word in the following sentence with the NER symbol. He also praised the decision of the al-Barnawi faction of Boko Haram to ease their struggle. --> \textcolor{blue}{\textbf{['(Delphin Kapaya,PER PER)', '(Oshine,Oshine)', '(Delphin,PER)']}} \\ \hline
\texttt{kin} &  Enter each word in the following sentence with the NER tag. Amazon plans to invest \$6.5 billion in online retail and will be able to access Bharti's low-cost Internet service if the two companies agree to the desired deal. --> (Amazon,B-ORG), (plan,O), (invest,O), (billion,O), (6,O), (O), (O), (5,O), (zamadorali,O), (in,O), (activity,O), (that), (buy,O), (true,O), (internet,O), (), (even,O), (price,O), (Ease,O), (O), (Want,O), (U), (internet), (U), (Bharti), (B-G), (G), (ORG), (O), (O), (O), (O), (O

Enter each word in the following sentence with the NER tag. They are known for their so-called Morning Worship where they sing a variety of songs from the book to help many. --> \textcolor{blue}{\textbf{[]}} \\ \hline
\texttt{lug} &  Each word in the following sentence has its own NER symbol. The sponsors gave me a scholarship to go to the United States to study for a master's degree and then the chicken business was given to my friend Geoffrey Lwanga who currently has over 7000 chickens. --> (People,O), (Give,O), (Help,O), (Give,O), (School,O), (Go,O), (In,O), (America,B-LOC), (Read,O), (Language,O), (Second,O), (Follow,O), (Business,O), (Chicken,O), (Ne,O), (Give,O), (Other), (Friend,O), (Off), (Geoff,B-PER), (Language,PER), (I), (O), (In), (Now

Each word in the following sentence has its own NER symbol. This is ours and we will raise funds to support her campaigns and to help her create posters and create works of art. Rose Namuli works for Pepsi and I gave her 2, 000. --> \textcolor{blue}{\textbf{['(America,LOC)', '(Other,Other)', '(Off,Off)']}} \\ \hline
 \bottomrule
\end{tabular}
\caption{Example few-shot prompts and their respective model outputs for the Translate adaptation method on masakhaNER. We use the same prompts for all models, but the reported outputs here are from one of the random seeds in the LLaMa2-7B experiments.}
\end{table*}

\begin{table*}[htbp]
\centering
\small
\begin{tabular}{p{0.2\columnwidth}p{1.8\columnwidth}}
\toprule
\textbf{Lang.}  & \textbf{Example Input + \textcolor{blue}{Output}} \\ \midrule
\texttt{bur} &  Mark each word with its NER tag in the following sentences: husband, B-PER, husband, I-PER, ((,I-PER), husband, I-PER, (), I-PER

Mark each word in the following sentence with its NER tag. Located in Alam, Yangon Province, and opened in 1963 by the Burmese timber industry. The courses that are open are: --> \textcolor{blue}{\textbf{['( (, PER)', '(,)', '( (, PER)', '(,)', '( (, PER)', '(,)']}} \\ \hline
\texttt{tha} &  Mark each word in the following sentences with the NER \# tag. (\#,O), (B-ORG), (A,I-ORG), (R,I-ORG), (B,I-ORG), (I,I-ORG), (N,I-ORG), (I,I-ORG), (T,I-ORG), (I,I-ORG) and (I,I-ORG) and then write the following:

Mark each word in the following sentences with the NER tag \# load \# load \# (logon \# A F 4) --> \textcolor{blue}{\textbf{['(logon\#AF4,logon\#AF4)', '(logon\#AF4,logon\#AF4)', '(logon\#AF4,logon\#AF4)', '(logon\#AF4,logon\#AF4)', '(logon\#AF4,logon\#AF4)', '(logon\#AF4,logon\#AF4)', '(logon\#AF4,logon\#AF4)', '(logon\#AF4,logon\#AF4)', '(logon\#AF4,logon\#AF4)', '(logon\#AF4,logon\#AF4)', '(logon\#AF4,logon\#AF4)', '(logon\#AF4,logon\#AF4)', '(logon\#AF4,logon\#AF4)', '(logon\#AF4,logon\#AF4)', '(logon\#AF4,logon\#AF4)', '(logon\#AF4,logon\#AF4)', '(logon\#AF4,logon\#AF4)', '(logon\#AF4,logon\#AF4)', '(logon\#AF4,logon\#AF4)', '(logon\#AF4,logon\#AF4)', '(logon\#AF4,logon\#AF4)', '(logon\#AF4,logon\#AF4)', '(logon\#AF4,logon\#AF4)', '(logon\#AF4,logon\#AF4)', '(logon\#AF4,logon\#AF4)']}} \\ \hline
 \bottomrule
\end{tabular}
\caption{Example few-shot prompts and their respective model outputs for the Translate adaptation method on wikiANN. We use the same prompts for all models, but the reported outputs here are from one of the random seeds in the LLaMa2-7B experiments.}
\end{table*}

\begin{table*}[htbp]
\centering
\small
\begin{tabular}{p{0.2\columnwidth}p{1.8\columnwidth}}
\toprule
\textbf{Lang.}  & \textbf{Example Input + \textcolor{blue}{Output}} \\ \midrule
\texttt{hau} &  Fitar da amsar karshe kawai ga tambayar lissafi. 
 Leah nada 32 chaculet, yar uwarta kuma 42.gudanawa suka rage musu? --> 39

Fitar da amsar karshe kawai ga tambayar lissafi. 
 Agwagin Janet suna yin kwai 16 a kullun. Tana yin karin kumallo da guda uku kowace safiya, sannan tana gasawa kawayenta guda hudu kullum. A kullum takan sayar da ragowar a kasuwar manoma akan dala 2 akan kowane kwai. Dala nawa take samu a kullum a kasuwar manoma? --> \textcolor{blue}{\textbf{39}} \\ \hline
\texttt{kin} &  Ibisohoka gusa igisubizo cyanyuma kubibazo byimibare. 
 Leah afite shokola 32 naho umuvandimwe we afite 42. Nibarya 35 bazaba basigaranye shokola zingahe zose hamwe? --> 39

Ibisohoka gusa igisubizo cyanyuma kubibazo byimibare. 
 Igishuhe cya Jane gitera amajyi 16 ku munsi,  buri mugitondo aryamo atatu kandi akora  umugati winshutiye akoresheje ane, agurisha asigaye mwisoko ryabahinzi buri munsi kugichiro cya 2 kuri buri jyi. Na ngahe mumadolali yinjiza ku munsi  mwisoko ryabahinzi ? --> \textcolor{blue}{\textbf{<NAN>}} \\ \hline
\texttt{lug} &  Fulumya eky’okuddamu ekisembayo kyokka ku kibuuzo ky’okubala. 
 Leah yalina kyokuleeti 32 ate nga muganda we ye yalina 42. Bwe baba nga baalyako 35, baasigazaawo kyokuleeti mmeka bombi omugatte? --> 39

Fulumya eky’okuddamu ekisembayo kyokka ku kibuuzo ky’okubala. 
 Embaata za Janet zibiika amagi 16 buli lunaku. Alya amagi asatu buli lunaku ku kyenkya n'afumbisa amalala ana g'ateeka mu bukkeeki bwa muffin bw'akolera mikwano gye. Agasigadde agatunda mu katale k'abalimi n'abalunzi buli lunaku nga buli ggi alitunda \$2. Afuna ssente mmeka buli lunaku mu katale k'abalimi n'abalunzi? --> \textcolor{blue}{\textbf{<NAN>}} \\ \hline
 \bottomrule
\end{tabular}
\caption{Example few-shot prompts and their respective model outputs for the LAFT adaptation method on AfriMGSM. We use the same prompts for all models, but the reported outputs here are from one of the random seeds in the LLaMa2-7B experiments.}
\end{table*}

\begin{table*}[htbp]
\centering
\small
\begin{tabular}{p{0.2\columnwidth}p{1.8\columnwidth}}
\toprule
\textbf{Lang.}  & \textbf{Example Input + \textcolor{blue}{Output}} \\ \midrule
\texttt{hau} &  Zabi zabin amsa daidai: A, B, C, ko D. 
 Wani masanin kimiyya ya auna dayamita na gashin mutum hudu. Dayamitocin, a ma'anin milimita, sune 0.091, 0.169, 0.17, da 0.023. Wanne in'ikwaliti ne ya kwatanta biyu daga dayamitocin biyu na gashin dan adam? 
 A: 0.17 > 0.023 
 B: 0.091 < 0.023 
 C: 0.169 > 0.17 
 D: 0.17 < 0.091 --> A

Zabi zabin amsa daidai: A, B, C, ko D. 
 Menene matsayin p a cikin 24 = 2p? 
 A: p = 4 
 B: p = 8 
 C: p = 12 
 D: p = 24 --> \textcolor{blue}{\textbf{B}} \\ \hline
\texttt{kin} &  Tora igisubizo gikwiye: A, B, C, cyangwa D. 
 Umuhanga yapimye diameter yimisatsi ine yabantu. Diameter, muri milimetero, yari 0.091, 0.169, 0.17, na 0.023. Ni ubuhe busumbane bugereranya neza diameter yimisatsi ibiri muriyo misatsi yabantu? 
 A: 0.17 > 0.023 
 B: 0.091 < 0.023 
 C: 0.169 > 0.17 
 D: 0.169 > 0.17 --> A

Tora igisubizo gikwiye: A, B, C, cyangwa D. 
 Nakahe gaciro ka p muri 24 = 2p?  
 A: p = 4 
 B: p = 8 
 C: p = 12 
 D: p = 24 --> \textcolor{blue}{\textbf{<NAN>}} \\ \hline
 \bottomrule
\end{tabular}
\caption{Example few-shot prompts and their respective model outputs for the LAFT adaptation method on AfriMMLU. We use the same prompts for all models, but the reported outputs here are from one of the random seeds in the LLaMa2-7B experiments.}
\end{table*}

\clearpage            % flush all pending floats
\includepdf[pages=1]{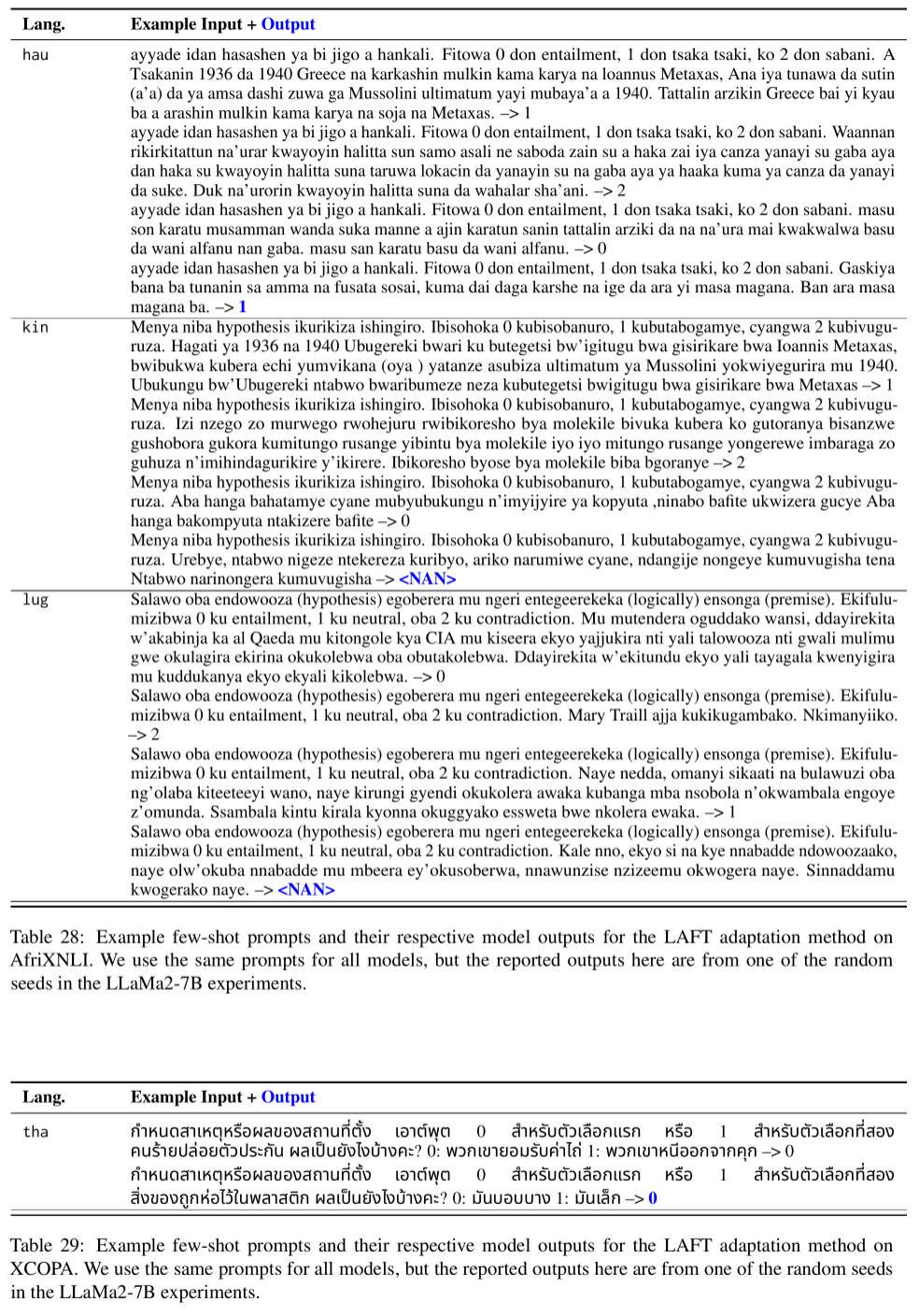}
\clearpage            % isolate this page from others

\clearpage            % flush all pending floats
\includepdf[pages=1]{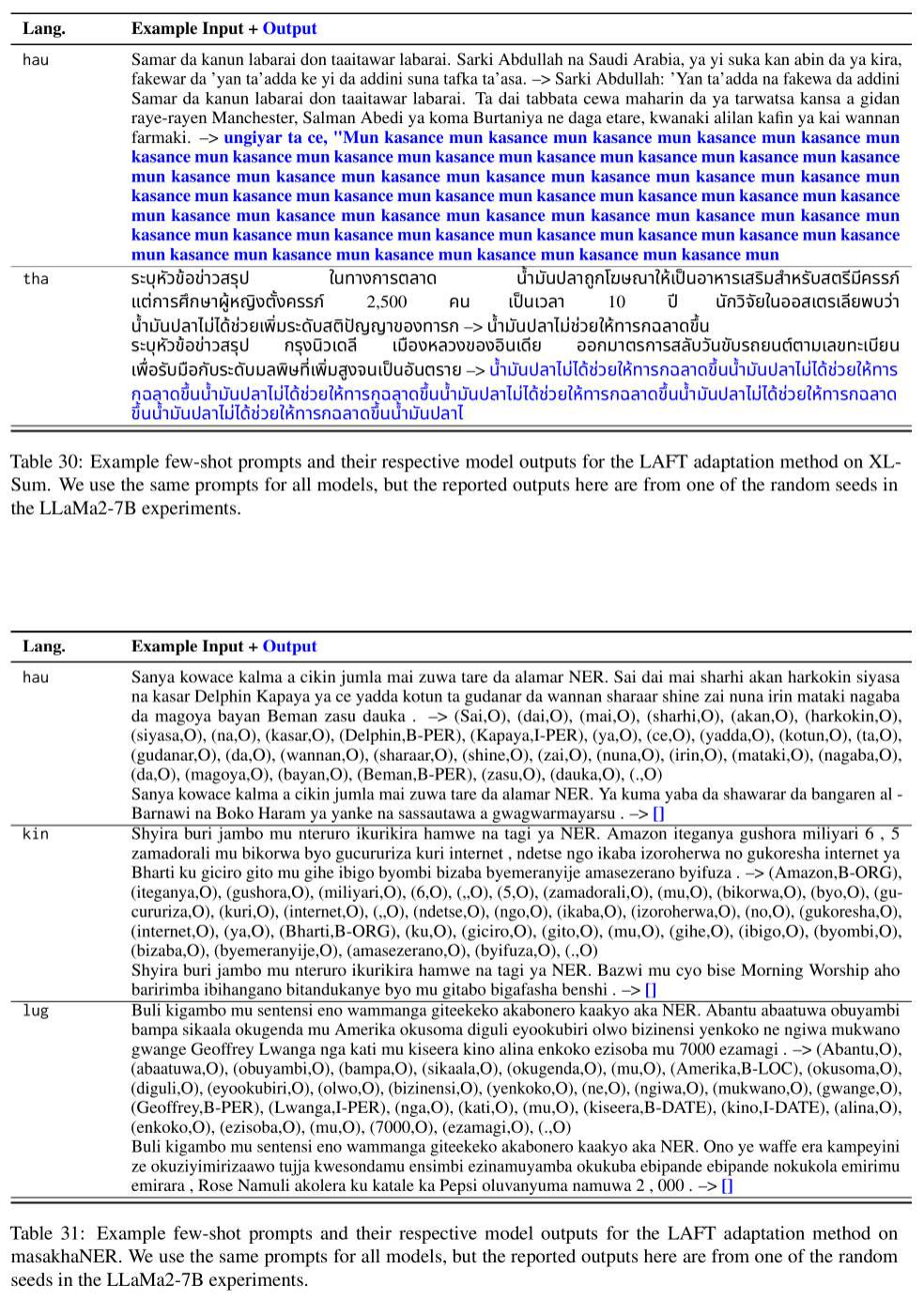}
\clearpage            % isolate this page from others

\setcounter{table}{31}
\begin{table*}[htbp]
\centering
\small
\begin{tabular}{p{0.2\columnwidth}p{1.8\columnwidth}}
\toprule
\textbf{Lang.}  & \textbf{Example Input + \textcolor{blue}{Output}} \\ \midrule
\texttt{hau} &  Fitar da amsar karshe kawai ga tambayar lissafi. 
 Leah nada 32 chaculet, yar uwarta kuma 42.gudanawa suka rage musu? --> 39

Fitar da amsar karshe kawai ga tambayar lissafi. 
 Agwagin Janet suna yin kwai 16 a kullun. Tana yin karin kumallo da guda uku kowace safiya, sannan tana gasawa kawayenta guda hudu kullum. A kullum takan sayar da ragowar a kasuwar manoma akan dala 2 akan kowane kwai. Dala nawa take samu a kullum a kasuwar manoma? --> \textcolor{blue}{\textbf{<NAN>}} \\ \hline
\texttt{kin} &  Ibisohoka gusa igisubizo cyanyuma kubibazo byimibare.  $<$unk$>$  Leah afite shokola  $<$unk$>$  naho umuvandimwe we afite  $<$unk$>$ . Nibarya  $<$unk$>$  bazaba basigaranye shokola zingahe zose hamwe? -->  $<$unk$>$ Ibisohoka gusa igisubizo cyanyuma kubibazo byimibare.  $<$unk$>$  Igishuhe cya Jane gitera amajyi  $<$unk$>$  ku munsi,  buri mugitondo aryamo atatu kandi akora  umugati winshutiye akoresheje ane, agurisha asigaye mwisoko ryabahinzi buri munsi kugichiro cya  $<$unk$>$  kuri buri jyi. Na ngahe mumadolali yinjiza ku munsi  mwisoko ryabahinzi ? --> \textcolor{blue}{\textbf{<NAN>}} \\ \hline
\texttt{lug} &  Fulumya eky’okuddamu ekisembayo kyokka ku kibuuzo ky’okubala.  $<$unk$>$  Leah yalina kyokuleeti  $<$unk$>$  ate nga muganda we ye yalina  $<$unk$>$ . Bwe baba nga baalyako  $<$unk$>$ , baasigazaawo kyokuleeti mmeka bombi omugatte? -- $<$unk$>$   $<$unk$>$ Fulumya eky’okuddamu ekisembayo kyokka ku kibuuzo ky’okubala.  $<$unk$>$  Embaata za Janet zibiika amagi  $<$unk$>$  buli lunaku. Alya amagi asatu buli lunaku ku kyenkya n'afumbisa amalala ana g'ateeka mu bukkeeki bwa muffin bw'akolera mikwano gye. Agasigadde agatunda mu katale k'abalimi n'abalunzi buli lunaku nga buli ggi alitunda  $<$unk$>$ . Afuna ssente mmeka buli lunaku mu katale k'abalimi n'abalunzi? -- $<$unk$>$  \textcolor{blue}{\textbf{<NAN>}} \\ \hline
 \bottomrule
\end{tabular}
\caption{Example few-shot prompts and their respective model outputs for the FOCUS adaptation method on AfriMGSM. We use the same prompts for all models, but the reported outputs here are from one of the random seeds in the LLaMa2-7B experiments.}
\end{table*}

\begin{table*}[htbp]
\centering
\small
\begin{tabular}{p{0.2\columnwidth}p{1.8\columnwidth}}
\toprule
\textbf{Lang.}  & \textbf{Example Input + \textcolor{blue}{Output}} \\ \midrule
\texttt{hau} &  Zabi zabin amsa daidai: A, B, C, ko D. 
 Wani masanin kimiyya ya auna dayamita na gashin mutum hudu. Dayamitocin, a ma'anin milimita, sune 0.091, 0.169, 0.17, da 0.023. Wanne in'ikwaliti ne ya kwatanta biyu daga dayamitocin biyu na gashin dan adam? 
 A: 0.17 > 0.023 
 B: 0.091 < 0.023 
 C: 0.169 > 0.17 
 D: 0.17 < 0.091 --> A

Zabi zabin amsa daidai: A, B, C, ko D. 
 Menene matsayin p a cikin 24 = 2p? 
 A: p = 4 
 B: p = 8 
 C: p = 12 
 D: p = 24 --> \textcolor{blue}{\textbf{A}} \\ \hline
\texttt{kin} &  Tora igisubizo gikwiye: A, B, C, cyangwa D.  $<$unk$>$  Umuhanga yapimye diameter yimisatsi ine yabantu. Diameter, muri milimetero, yari  $<$unk$>$ . $<$unk$>$ ,  $<$unk$>$ . $<$unk$>$ ,  $<$unk$>$ . $<$unk$>$ , na  $<$unk$>$ . $<$unk$>$ . Ni ubuhe busumbane bugereranya neza diameter yimisatsi ibiri muriyo misatsi yabantu?  $<$unk$>$  A:  $<$unk$>$ . $<$unk$>$  >  $<$unk$>$ . $<$unk$>$   $<$unk$>$  B:  $<$unk$>$ . $<$unk$>$  <  $<$unk$>$ . $<$unk$>$   $<$unk$>$  C:  $<$unk$>$ . $<$unk$>$  >  $<$unk$>$ . $<$unk$>$   $<$unk$>$  D:  $<$unk$>$ . $<$unk$>$  >  $<$unk$>$ . $<$unk$>$  --> A $<$unk$>$ Tora igisubizo gikwiye: A, B, C, cyangwa D.  $<$unk$>$  Nakahe gaciro ka p muri  $<$unk$>$  =  $<$unk$>$ p?   $<$unk$>$  A: p =  $<$unk$>$   $<$unk$>$  B: p =  $<$unk$>$   $<$unk$>$  C: p =  $<$unk$>$   $<$unk$>$  D: p =  $<$unk$>$  --> \textcolor{blue}{\textbf{<NAN>}} \\ \hline
\texttt{lug} &  Londa eky’okuddamu ekituufu: A, B, C, oba D.  $<$unk$>$  Munnassaayansi yapima obugazi bw'enviiri z'omuntu nnya. Obugazi mu butundutundu buli  $<$unk$>$ . $<$unk$>$ ,  $<$unk$>$ . $<$unk$>$ ,  $<$unk$>$ . $<$unk$>$ , ne  $<$unk$>$ . $<$unk$>$ . Bukwatane ki wakati w'emiwendo egitenkanankana egisobola okukozesebwa mu butuufu okugeraageranya obugazi bw'enviiri z'omuntu bbiri?  $<$unk$>$  A:  $<$unk$>$ . $<$unk$>$   $<$unk$>$   $<$unk$>$ . $<$unk$>$   $<$unk$>$  B:  $<$unk$>$ . $<$unk$>$   $<$unk$>$   $<$unk$>$ . $<$unk$>$   $<$unk$>$  C:  $<$unk$>$ . $<$unk$>$   $<$unk$>$   $<$unk$>$ . $<$unk$>$   $<$unk$>$  D:  $<$unk$>$ . $<$unk$>$   $<$unk$>$   $<$unk$>$ . $<$unk$>$  -- $<$unk$>$  A $<$unk$>$ Londa eky’okuddamu ekituufu: A, B, C, oba D.  $<$unk$>$  p erina muwendo ki mu  $<$unk$>$   $<$unk$>$   $<$unk$>$ p?  $<$unk$>$  A: p  $<$unk$>$   $<$unk$>$   $<$unk$>$  B: p  $<$unk$>$   $<$unk$>$   $<$unk$>$  C: p  $<$unk$>$   $<$unk$>$   $<$unk$>$  D: p  $<$unk$>$   $<$unk$>$  -- $<$unk$>$  \textcolor{blue}{\textbf{<NAN>}} \\ \hline
 \bottomrule
\end{tabular}
\caption{Example few-shot prompts and their respective model outputs for the FOCUS adaptation method on AfriMMLU. We use the same prompts for all models, but the reported outputs here are from one of the random seeds in the LLaMa2-7B experiments.}
\end{table*}

\begin{table*}[htbp]
\centering
\small
\begin{tabular}{p{0.2\columnwidth}p{1.8\columnwidth}}
\toprule
\textbf{Lang.}  & \textbf{Example Input + \textcolor{blue}{Output}} \\ \midrule
\texttt{hau} &  kayyade idan hasashen ya bi jigo a hankali. Fitowa 0 don entailment, 1 don tsaka tsaki, ko 2 don sabani. 
 A Tsakanin 1936 da 1940 Greece na karkashin mulkin kama karya na loannus Metaxas, Ana iya tunawa da sutin (a'a)  da ya amsa dashi zuwa ga Mussolini ultimatum yayi mubaya'a a 1940. Tattalin arzikin Greece bai yi kyau ba a karkashin mulkin kama karya na soja na Metaxas. --> 1

kayyade idan hasashen ya bi jigo a hankali. Fitowa 0 don entailment, 1 don tsaka tsaki, ko 2 don sabani. 
 Wadannan rikirkitattun na'urar kwayoyin halitta sun samo asali ne saboda zabin su a haka zai iya canza yanayi su gaba daya dan haka su kwayoyin halitta suna taruwa lokacin da yanayin su na gaba daya ya habaka kuma ya canza da yanayi da suke. Duk na'urorin kwayoyin halitta suna da wahalar sha'ani. --> 2

kayyade idan hasashen ya bi jigo a hankali. Fitowa 0 don entailment, 1 don tsaka tsaki, ko 2 don sabani. 
 masu son karatu musamman wanda suka manne a ajin karatun sanin tattalin arziki da na na'ura mai kwakwalwa basu da wani alfanu nan gaba. masu san karatu basu da wani alfanu. --> 0

kayyade idan hasashen ya bi jigo a hankali. Fitowa 0 don entailment, 1 don tsaka tsaki, ko 2 don sabani. 
 Gaskiya bana ba tunanin sa amma na fusata sosai, kuma dai daga karshe na bige da kara yi masa magana. Ban kara masa magana ba. --> \textcolor{blue}{\textbf{<NAN>}} \\ \hline
\texttt{kin} &  Menya niba hypothesis ikurikiza ishingiro. Ibisohoka  $<$unk$>$  kubisobanuro,  $<$unk$>$  kubutabogamye, cyangwa  $<$unk$>$  kubivuguruza.  $<$unk$>$  Hagati ya  $<$unk$>$  na  $<$unk$>$  Ubugereki bwari ku butegetsi bw'igitugu bwa gisirikare bwa Ioannis Metaxas, bwibukwa kubera echi yumvikana (oya ) yatanze asubiza ultimatum ya Mussolini yokwiyegurira mu  $<$unk$>$ . Ubukungu bw'Ubugereki ntabwo bwaribumeze neza kubutegetsi bwigitugu bwa gisirikare bwa Metaxas -->  $<$unk$>$ Menya niba hypothesis ikurikiza ishingiro. Ibisohoka  $<$unk$>$  kubisobanuro,  $<$unk$>$  kubutabogamye, cyangwa  $<$unk$>$  kubivuguruza.  $<$unk$>$  Izi nzego zo murwego rwohejuru rwibikoresho bya molekile bivuka kubera ko gutoranya bisanzwe gushobora gukora kumitungo rusange yibintu bya molekile iyo iyo mitungo rusange yongerewe imbaraga zo guhuza n'imihindagurikire y'ikirere. $<$unk$>$  Ibikoresho byose bya molekile biba bgoranye -->  $<$unk$>$ Menya niba hypothesis ikurikiza ishingiro. Ibisohoka  $<$unk$>$  kubisobanuro,  $<$unk$>$  kubutabogamye, cyangwa  $<$unk$>$  kubivuguruza.  $<$unk$>$  Aba hanga bahatamye cyane mubyubukungu n'imyijyire ya kopyuta ,ninabo bafite ukwizera gucye Aba hanga bakompyuta ntakizere bafite -->  $<$unk$>$ Menya niba hypothesis ikurikiza ishingiro. Ibisohoka  $<$unk$>$  kubisobanuro,  $<$unk$>$  kubutabogamye, cyangwa  $<$unk$>$  kubivuguruza.  $<$unk$>$  Urebye, ntabwo nigeze ntekereza kuribyo, ariko narumiwe cyane, ndangije nongeye kumuvugisha tena Ntabwo narinongera kumuvugisha  --> \textcolor{blue}{\textbf{<NAN>}} \\ \hline
\texttt{lug} &  Salawo oba endowooza (hypothesis) egoberera mu ngeri entegeerekeka (logically) ensonga (premise). Ekifulumizibwa  $<$unk$>$  ku entailment,  $<$unk$>$  ku neutral, oba  $<$unk$>$  ku contradiction.  $<$unk$>$  Wakati wa  $<$unk$>$  ne  $<$unk$>$  Greece yali wansi w'obufuzi obwa nnakyemalira  Ioannis Metaxas, ajjukirwa ennyo olw'enziramu ya 'Nedda' gye yayanukula Mussolini bwe yali amuwadde nsalessale w'okuwanika nga awanguddwa mu  $<$unk$>$ . Ebyenfuna bya Greece tebyatambula bulungi mu kiseera ng'eri wansi wa nnaakyemalira Metaxas. -- $<$unk$>$   $<$unk$>$ Salawo oba endowooza (hypothesis) egoberera mu ngeri entegeerekeka (logically) ensonga (premise). Ekifulumizibwa  $<$unk$>$  ku entailment,  $<$unk$>$  ku neutral, oba  $<$unk$>$  ku contradiction.  $<$unk$>$  Obusimu obw'eddaala erya waggulu busituka kubanga obutonde bubeera n'obusobozi okukikola bwe bumanyiira okukikola. Obusimu bwonna bwa ddaala lya waggulu. -- $<$unk$>$   $<$unk$>$ Salawo oba endowooza (hypothesis) egoberera mu ngeri entegeerekeka (logically) ensonga (premise). Ekifulumizibwa  $<$unk$>$  ku entailment,  $<$unk$>$  ku neutral, oba  $<$unk$>$  ku contradiction.  $<$unk$>$  Abantu abatalina bukugu mu kubeerana na balala abanoonya obubudamu mu bibiina by'amasomo g'ebyenfuna oba ebya kompyuta, ate bo basingawo mu kubeera mu mbeera y'obutaba na ssuubi. Abantu abatalina bukugu mu kubeerana na balala tebalina ssuubi. -- $<$unk$>$   $<$unk$>$ Salawo oba endowooza (hypothesis) egoberera mu ngeri entegeerekeka (logically) ensonga (premise). Ekifulumizibwa  $<$unk$>$  ku entailment,  $<$unk$>$  ku neutral, oba  $<$unk$>$  ku contradiction.  $<$unk$>$  Kale nno, ekyo si na kye nnabadde ndowoozaako, naye olw'okuba nnabadde mu mbeera ey'okusoberwa, nnawunzise nzizeemu okwogera naye. Sinnaddamu kwogerako naye. -- $<$unk$>$  \textcolor{blue}{\textbf{<NAN>}} \\ \hline
 \bottomrule
\end{tabular}
\caption{Example few-shot prompts and their respective model outputs for the FOCUS adaptation method on AfriXNLI. We use the same prompts for all models, but the reported outputs here are from one of the random seeds in the LLaMa2-7B experiments.}
\end{table*}

\clearpage            % flush all pending floats
\includepdf[pages=1]{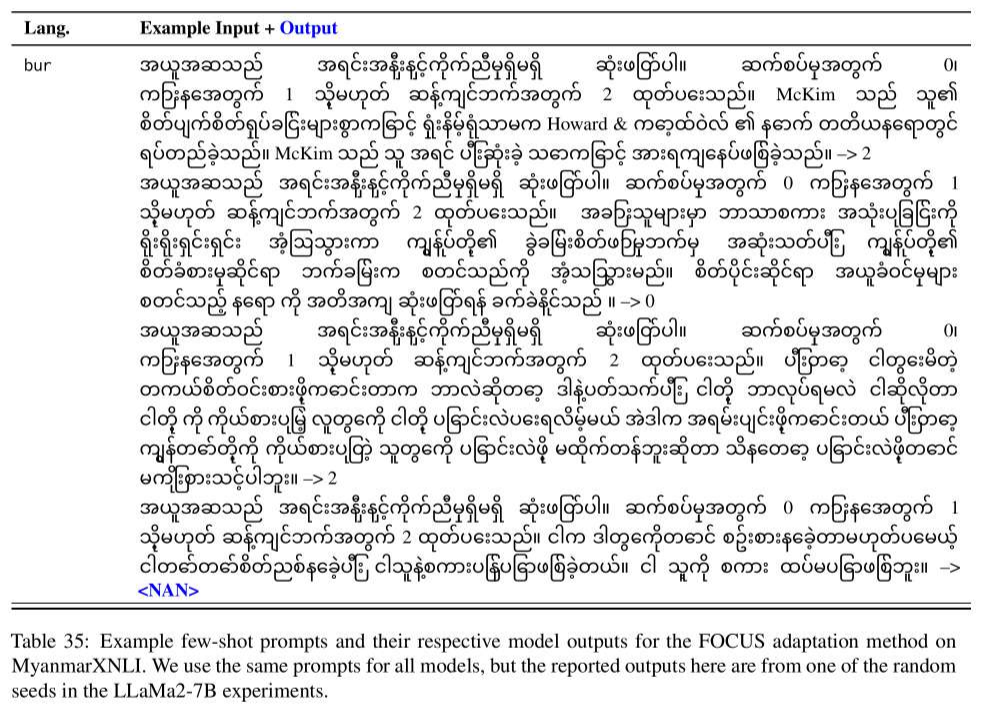}
\clearpage            % isolate this page from others

\clearpage            % flush all pending floats
\includepdf[pages=1]{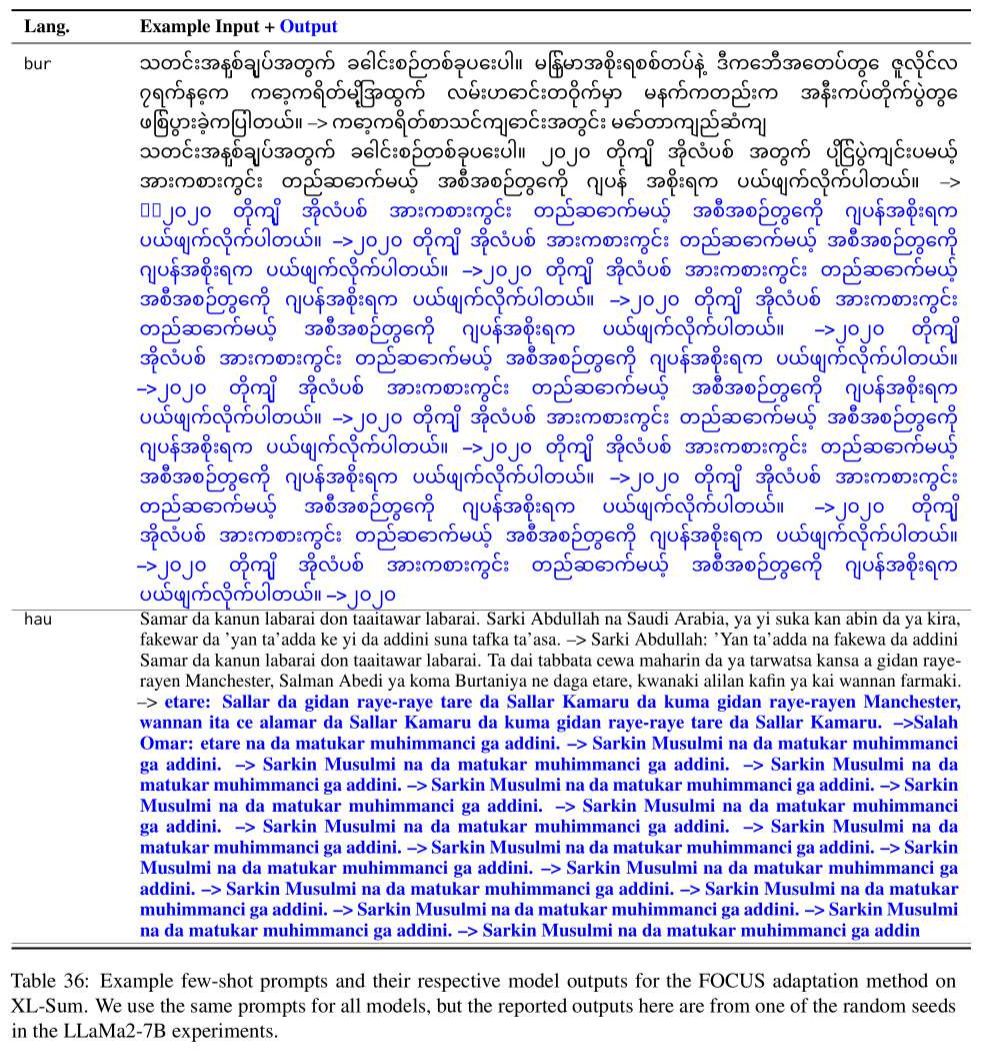}
\clearpage            % isolate this page from others

\clearpage            % flush all pending floats
\includepdf[pages=1]{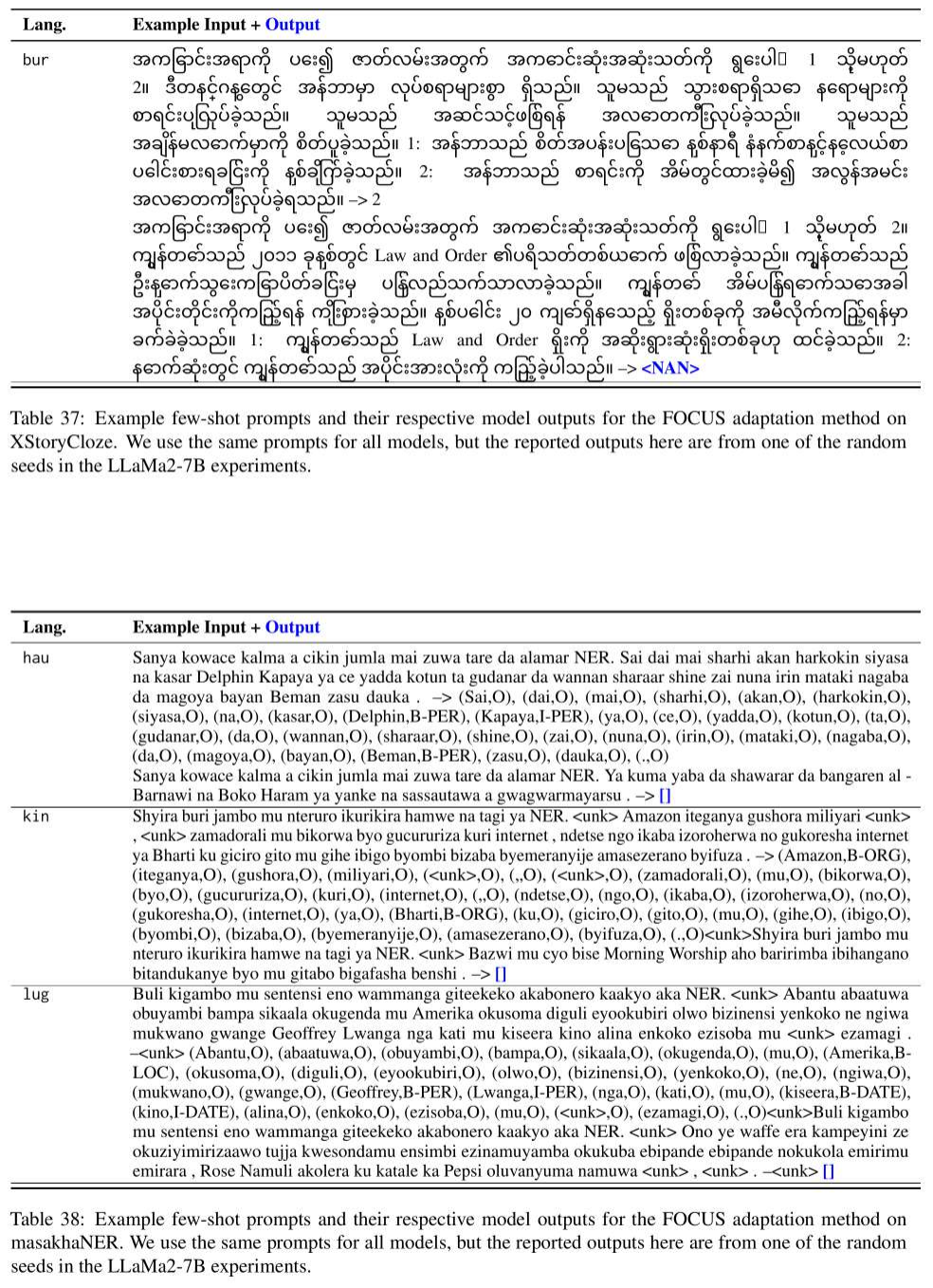}
\clearpage            % isolate this page from others

\clearpage            % flush all pending floats
\includepdf[pages=1]{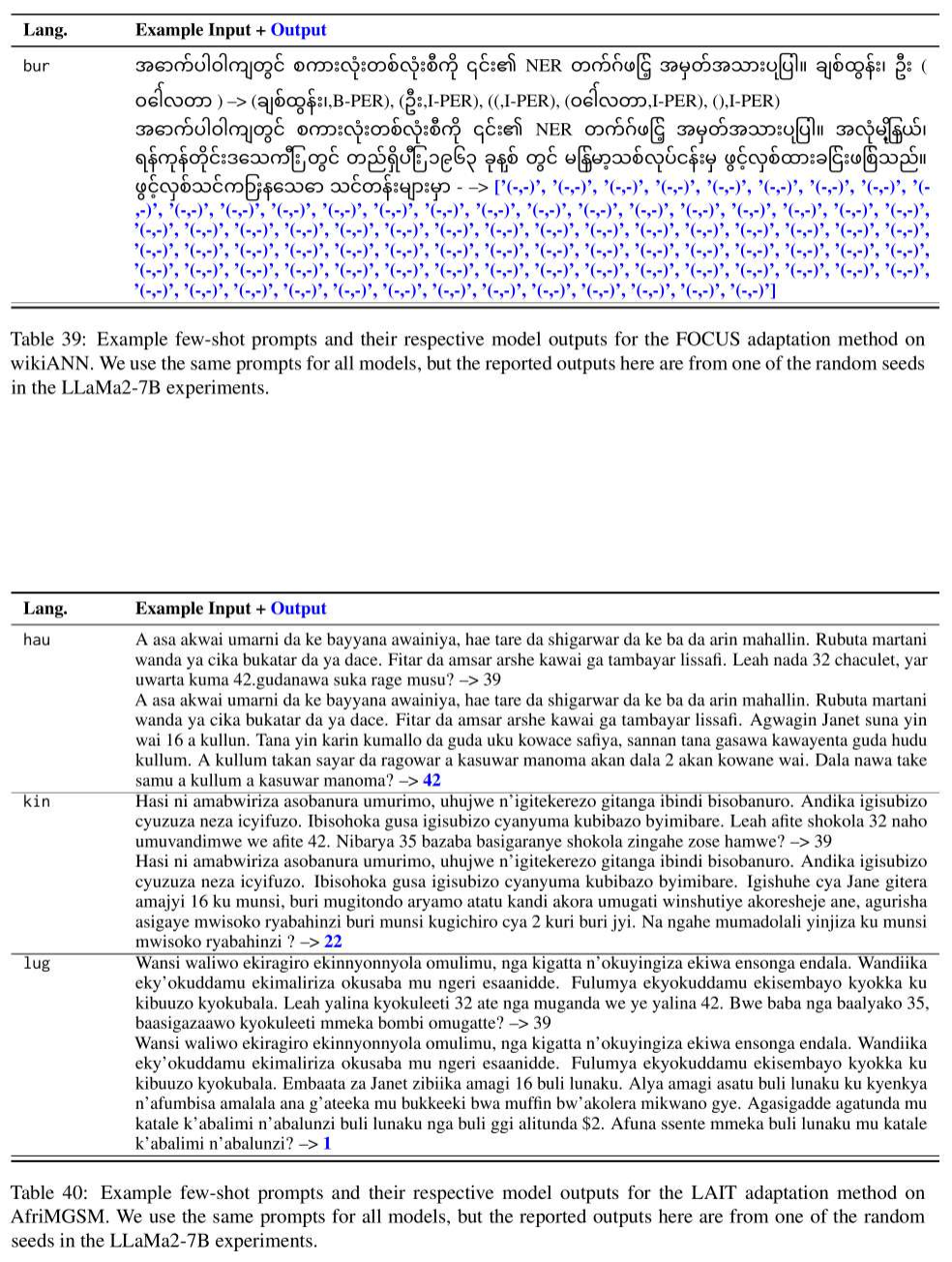}
\clearpage            % isolate this page from others

\clearpage            % flush all pending floats
\includepdf[pages=1]{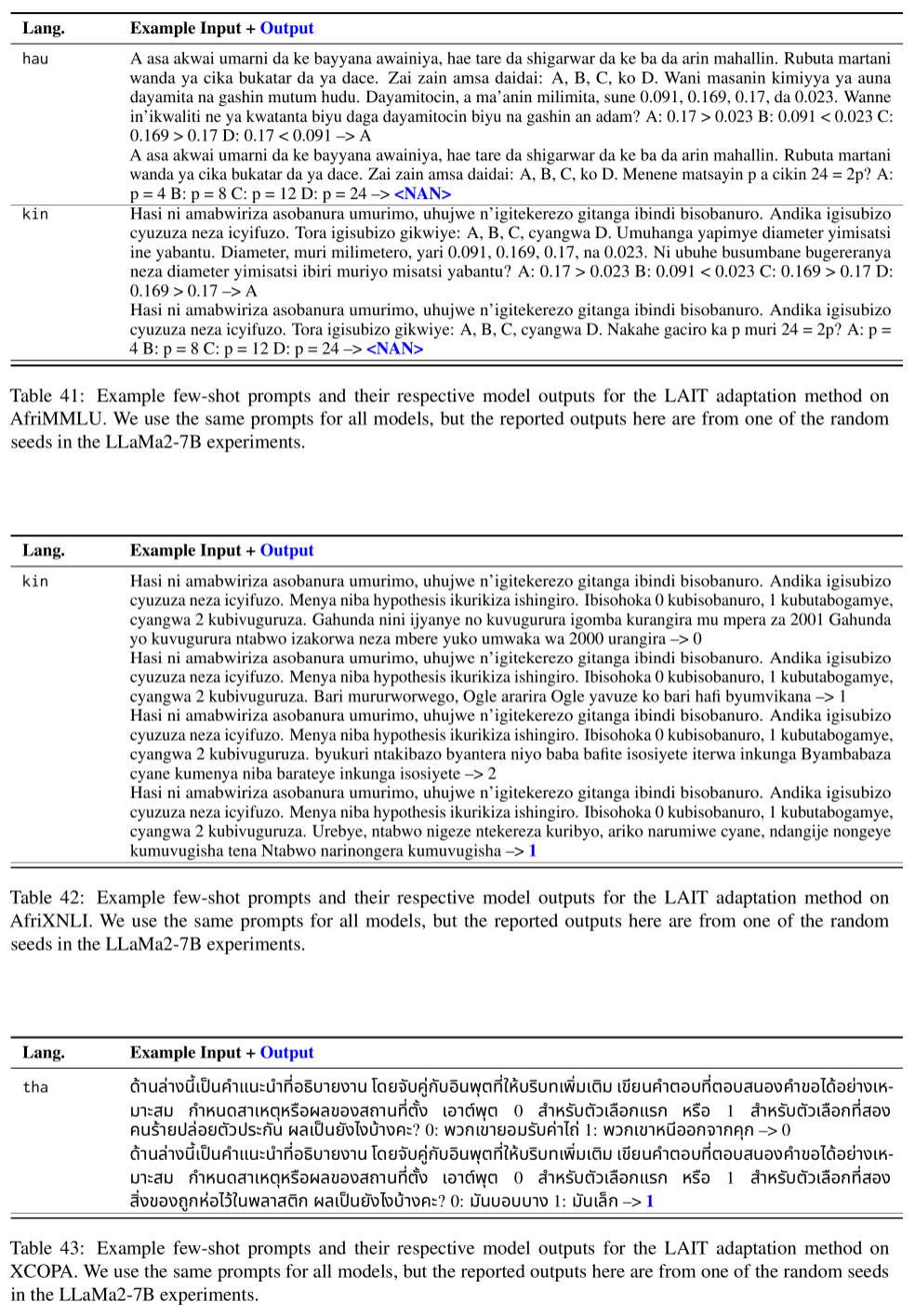}
\clearpage            % isolate this page from others

\setcounter{table}{43}
\begin{table*}[htbp]
\centering
\small
\begin{tabular}{p{0.2\columnwidth}p{1.8\columnwidth}}
\toprule
\textbf{Lang.}  & \textbf{Example Input + \textcolor{blue}{Output}} \\ \midrule
\texttt{hau} &  A kasa akwai umarni da ke bayyana dawainiya, hade tare da shigarwar da ke ba da karin mahallin. Rubuta martani wanda ya cika bukatar da ya dace. 
Samar da kanun labarai don takaitawar labarai. 
 Sarki Abdullah na Saudi Arabia, ya yi suka kan abin da ya kira, fakewar da 'yan ta'adda ke yi da addini suna tafka ta'asa. --> Sarki Abdullah: 'Yan ta'adda na fakewa da addini

A kasa akwai umarni da ke bayyana dawainiya, hade tare da shigarwar da ke ba da karin mahallin. Rubuta martani wanda ya cika bukatar da ya dace. 
Samar da kanun labarai don takaitawar labarai. 
 Ta dai tabbata cewa maharin da ya tarwatsa kansa a gidan raye-rayen Manchester, Salman Abedi ya koma Burtaniya ne daga ketare, kwanaki kalilan kafin ya kai wannan farmaki. --> \textcolor{blue}{\textbf{daya daga cikin dabarun talla, kamar yadda suke yin tafiya a cikin dabarun talla. $<$unk$>$  $<$unk$>$  $<$unk$>$  $<$unk$>$  $<$unk$>$  $<$unk$>$  $<$unk$>$  $<$unk$>$  $<$unk$>$  $<$unk$>$  $<$unk$>$  $<$unk$>$  $<$unk$>$  $<$unk$>$  $<$unk$>$ \newline  $<$unk$>$  $<$unk$>$  $<$unk$>$  $<$unk$>$  $<$unk$>$  $<$unk$>$  $<$unk$>$  $<$unk$>$  $<$unk$>$  $<$unk$>$  $<$unk$>$  $<$unk$>$  $<$unk$>$  $<$unk$>$  $<$unk$>$ \newline $<$unk$>$  $<$unk$>$  $<$unk$>$  $<$unk$>$  $<$unk$>$  $<$unk$>$  $<$unk$>$  $<$unk$>$  $<$unk$>$  $<$unk$>$  $<$unk$>$  $<$unk$>$  $<$unk$>$  $<$unk$>$  $<$unk$>$ \newline $<$unk$>$  $<$unk$>$  $<$unk$>$  $<$unk$>$  $<$unk$>$  $<$unk$>$  $<$unk$>$  $<$unk$>$  $<$unk$>$  $<$unk$>$  $<$unk$>$  $<$unk$>$  $<$unk$>$  $<$unk$>$  $<$unk$>$ \newline $<$unk$>$  $<$unk$>$  $<$unk$>$  $<$unk$>$  $<$unk$>$  $<$unk$>$  $<$unk$>$  $<$unk$>$  $<$unk$>$  $<$unk$>$ \ $<$unk$>$  $<$unk$>$  $<$unk$>$  $<$unk$>$  $<$unk$>$ \newline $<$unk$>$  $<$unk$>$  $<$unk$>$  $<$unk$>$  $<$unk$>$  $<$unk$>$  $<$unk$>$  $<$unk$>$  $<$unk$>$  $<$unk$>$  $<$unk$>$  $<$unk$>$  $<$unk$>$  $<$unk$>$  $<$unk$>$ \newline $<$unk$>$  $<$unk$>$  $<$unk$>$  $<$unk$>$  $<$unk$>$  $<$unk$>$  $<$unk$>$  $<$unk$>$  $<$unk$>$  $<$unk$>$  $<$unk$>$  $<$unk$>$  $<$unk$>$  $<$unk$>$  $<$unk$>$ \newline $<$unk$>$  $<$unk$>$  $<$unk$>$  $<$unk$>$  $<$unk$>$  $<$unk$>$  $<$unk$>$  $<$unk$>$  $<$unk$>$  $<$unk$>$  $<$unk$>$  $<$unk$>$  $<$unk$>$  $<$unk$>$  $<$unk$>$ \newline $<$unk$>$  $<$unk$>$  $<$unk$>$  $<$unk$>$  $<$unk$>$  $<$unk$>$  $<$unk$>$  $<$unk$>$  $<$unk$>$  $<$unk$>$  $<$unk$>$  $<$unk$>$  $<$unk$>$  $<$unk$>$  $<$unk$>$ \newline $<$unk$>$  $<$unk$>$  $<$unk$>$  $<$unk$>$  $<$unk$>$  $<$unk$>$  $<$unk$>$  $<$unk$>$  $<$unk$>$  $<$unk$>$  $<$unk$>$  $<$unk$>$  $<$unk$>$  $<$unk$>$  $<$unk$>$ \newline $<$unk$>$  $<$unk$>$  $<$unk$>$  $<$unk$>$  $<$unk$>$  $<$unk$>$  $<$unk$>$  $<$unk$>$  $<$unk$>$  $<$unk$>$  $<$unk$>$  $<$unk$>$  $<$unk$>$  $<$unk$>$  $<$unk$>$ \newline $<$unk$>$  $<$unk$>$  $<$unk$>$  $<$unk$>$  $<$unk$>$  $<$unk$>$  $<$unk$>$  $<$unk$>$  $<$unk$>$  $<$unk$>$  $<$unk$>$  $<$unk$>$  $<$unk$>$  $<$unk$>$  $<$unk$>$ \newline $<$unk$>$  $<$unk$>$  $<$unk$>$  $<$unk$>$  $<$unk$>$  $<$unk$>$  $<$unk$>$  $<$unk$>$  $<$unk$>$  $<$unk$>$  $<$unk$>$  $<$unk$>$  $<$unk$>$  $<$unk$>$  $<$unk$>$ \newline $<$unk$>$  $<$unk$>$  $<$unk$>$  $<$unk$>$  $<$unk$>$  $<$unk$>$  $<$unk$>$  $<$unk$>$  $<$unk$>$  $<$unk$>$  $<$unk$>$  $<$unk$>$  $<$unk$>$  $<$unk$>$  $<$unk$>$ \newline $<$unk$>$  $<$unk$>$  $<$unk$>$  $<$unk$>$  $<$unk$>$  $<$unk$>$ }} \\ \hline
 \bottomrule
\end{tabular}
\caption{Example few-shot prompts and their respective model outputs for the LAIT adaptation method on XL-Sum. We use the same prompts for all models, but the reported outputs here are from one of the random seeds in the LLaMa2-7B experiments.}
\end{table*}

\begin{table*}[htbp]
\centering
\small
\begin{tabular}{p{0.2\columnwidth}p{1.8\columnwidth}}
\toprule
\textbf{Lang.}  & \textbf{Example Input + \textcolor{blue}{Output}} \\ \midrule
\texttt{hau} &  A kasa akwai umarni da ke bayyana dawainiya, hade tare da shigarwar da ke ba da karin mahallin. Rubuta martani wanda ya cika bukatar da ya dace. 
Sanya kowace kalma a cikin jumla mai zuwa tare da alamar NER. 
 Sai dai mai sharhi akan harkokin siyasa na kasar Delphin Kapaya ya ce yadda kotun ta gudanar da wannan shara’ar shine zai nuna irin mataki nagaba da magoya bayan Beman zasu dauka . --> (Sai,O), (dai,O), (mai,O), (sharhi,O), (akan,O), (harkokin,O), (siyasa,O), (na,O), (kasar,O), (Delphin,B-PER), (Kapaya,I-PER), (ya,O), (ce,O), (yadda,O), (kotun,O), (ta,O), (gudanar,O), (da,O), (wannan,O), (shara’ar,O), (shine,O), (zai,O), (nuna,O), (irin,O), (mataki,O), (nagaba,O), (da,O), (magoya,O), (bayan,O), (Beman,B-PER), (zasu,O), (dauka,O), (.,O)

A kasa akwai umarni da ke bayyana dawainiya, hade tare da shigarwar da ke ba da karin mahallin. Rubuta martani wanda ya cika bukatar da ya dace. 
Sanya kowace kalma a cikin jumla mai zuwa tare da alamar NER. 
 Ya kuma yaba da shawarar da bangaren al - Barnawi na Boko Haram ya yanke na sassautawa a gwagwarmayarsu . --> \textcolor{blue}{\textbf{[]}} \\ \hline
\texttt{kin} &  Hasi ni amabwiriza asobanura umurimo, uhujwe n'igitekerezo gitanga ibindi bisobanuro. Andika igisubizo cyuzuza neza icyifuzo. 
 Shyira buri jambo mu nteruro ikurikira hamwe na tagi ya NER. 
 Amazon iteganya gushora miliyari 6 , 5 z’amadorali mu bikorwa byo gucururiza kuri internet , ndetse ngo ikaba izoroherwa no gukoresha internet ya Bharti ku giciro gito mu gihe ibigo byombi bizaba byemeranyije amasezerano byifuza . --> (Amazon,B-ORG), (iteganya,O), (gushora,O), (miliyari,O), (6,O), (,,O), (5,O), (z’amadorali,O), (mu,O), (bikorwa,O), (byo,O), (gucururiza,O), (kuri,O), (internet,O), (,,O), (ndetse,O), (ngo,O), (ikaba,O), (izoroherwa,O), (no,O), (gukoresha,O), (internet,O), (ya,O), (Bharti,B-ORG), (ku,O), (giciro,O), (gito,O), (mu,O), (gihe,O), (ibigo,O), (byombi,O), (bizaba,O), (byemeranyije,O), (amasezerano,O), (byifuza,O), (.,O)

Hasi ni amabwiriza asobanura umurimo, uhujwe n'igitekerezo gitanga ibindi bisobanuro. Andika igisubizo cyuzuza neza icyifuzo. 
 Shyira buri jambo mu nteruro ikurikira hamwe na tagi ya NER. 
 Bazwi mu cyo bise ‘Morning Worship’ aho baririmba ibihangano bitandukanye byo mu gitabo bigafasha benshi . --> \textcolor{blue}{\textbf{[]}} \\ \hline
\texttt{lug} &  Wansi waliwo ekiragiro ekinnyonnyola omulimu, nga kigatta n'okuyingiza ekiwa ensonga endala. Wandiika eky'okuddamu ekimaliriza okusaba mu ngeri esaanidde. 
Buli kigambo mu sentensi eno wammanga giteekeko akabonero kaakyo aka NER. 
 Abantu abaatuwa obuyambi bampa sikaala okugenda mu Amerika okusoma diguli eyookubiri olwo bizinensi yenkoko ne ngiwa mukwano gwange Geoffrey Lwanga nga kati mu kiseera kino alina enkoko ezisoba mu 7000 ezamagi . --> (Abantu,O), (abaatuwa,O), (obuyambi,O), (bampa,O), (sikaala,O), (okugenda,O), (mu,O), (Amerika,B-LOC), (okusoma,O), (diguli,O), (eyookubiri,O), (olwo,O), (bizinensi,O), (yenkoko,O), (ne,O), (ngiwa,O), (mukwano,O), (gwange,O), (Geoffrey,B-PER), (Lwanga,I-PER), (nga,O), (kati,O), (mu,O), (kiseera,B-DATE), (kino,I-DATE), (alina,O), (enkoko,O), (ezisoba,O), (mu,O), (7000,O), (ezamagi,O), (.,O)

Wansi waliwo ekiragiro ekinnyonnyola omulimu, nga kigatta n'okuyingiza ekiwa ensonga endala. Wandiika eky'okuddamu ekimaliriza okusaba mu ngeri esaanidde. 
Buli kigambo mu sentensi eno wammanga giteekeko akabonero kaakyo aka NER. 
 Ono ye waffe era kampeyini ze okuziyimirizaawo tujja kwesondamu ensimbi ezinamuyamba okukuba ebipande ebipande nokukola emirimu emirara , Rose Namuli akolera ku katale ka Pepsi oluvanyuma namuwa 2 , 000 . --> \textcolor{blue}{\textbf{[]}} \\ \hline
 \bottomrule
\end{tabular}
\caption{Example few-shot prompts and their respective model outputs for the LAIT adaptation method on masakhaNER. We use the same prompts for all models, but the reported outputs here are from one of the random seeds in the LLaMa2-7B experiments.}
\end{table*}

\end{document}